%% file: root.tex
\documentclass[letterpaper, 10 pt, conference]{ieeeconf}  

\IEEEoverridecommandlockouts                              

\usepackage{graphicx} 
\usepackage{mathptmx} 
\usepackage{amsmath} 
\usepackage{amssymb}  

\usepackage{multirow}
\usepackage{tabularx}

\usepackage{caption}
\usepackage{subcaption}
\usepackage{hyperref}

\usepackage{float}

\title{\LARGE \bf
U-Net for MAV-based Penstock Inspection: an Investigation of Focal Loss in Multi-class Segmentation for Corrosion Identification 
}

\author{Ty Nguyen$^{1}$, Tolga \"Ozaslan $^{1}$, Ian D. Miller$^{1}$, James Keller$^{1}$,  
Giuseppe Loianno$^{1}$, Camillo J. Taylor$^{1}$ \\ Daniel D. Lee$^{1}$, Vijay Kumar$^{1}$, Joseph H. Harwood$^{2}$, Jennifer Wozencraft$^{2}$
\thanks{This work was supported by the MAST Collaborative Technology Alliance - Contract No. W911NF-08-2-0004, ARL grant W911NF-08-2-0004, ONR grants N00014-07-1-0829, N00014-14-1-0510, ARO grant W911NF-13-1-0350, NSF grants IIS-1426840, IIS-1138847, DARPA grants HR001151626, HR0011516850}
\thanks{
T. \"Ozaslan acknowledges the fellowship from The Republic of Turkey
Ministry of National Education.}
\thanks{$^{1}$are with the
GRASP Lab, University of Pennsylvania, Philadelphia, PA 19104 USA (email: \{
        {\tt\small{tynguyen, ozaslan, iandm, jfkeller, cjtaylor, kumar}\}@seas.upenn.edu}}%
\thanks{$^{2}$ are with the United States Army Corps of Engineers, Washington, DC 20314 USA (email: \{
        {\tt\small{joseph.h.harwood, jennifer.m.wozencraft}\}@usace.army.mil}}%
}

\begin{document}

\maketitle
\thispagestyle{empty}
\pagestyle{empty}

\begin{abstract}
Periodical inspection and maintenance of critical infrastructure such as dams, penstocks and locks are of significant importance to prevent catastrophic failures.
Conventional manual inspection methods require inspectors to climb along a penstock to spot corrosion, rust and crack formation which is unsafe, labor-intensive, and requires intensive training.
This work presents an alternative approach using a Micro Aerial Vehicle (MAV) that autonomously flies to collect imagery which is then fed into a pretrained deep-learning model to identify corrosion. 
Our simplified U-Net trained with less than $40$ image samples can do inference at $12$ fps on a single GPU. 
We analyze different loss functions to solve the class imbalance problem, followed by a discussion on choosing proper metrics and weights for object classes. 
Results obtained with the dataset collected from Center Hill Dam, TN show that focal loss function, combined with a proper set of class weights yield better segmentation results than the base loss, Softmax cross entropy. 
Our method can be used in combination with \cite{ozaslan2017autonomous} to offer a complete, safe and cost-efficient solution to autonomous infrastructure inspection.  
\end{abstract}

\input{tex/introduction_tolga}
\input{tex/related_work_tolga}
\input{tex/dataset}
\input{tex/problem}
\input{tex/class_imbalance}
\input{tex/training}
\input{tex/evaluation}
\input{tex/conclusion}

\input{tex/appendix}
\bibliographystyle{bib/IEEEtran}
\bibliography{bib/bib.bib}

\end{document}

%% file: tex/introduction_tolga.tex
\section{Introduction}
\label{sec:introduction}

Deep learning has become the \textit{de-facto} approach in image and speech recognition, text processing, multi-modal learning with superior accuracy and robustness over other machine learning approaches \cite{lecun.nature15, deng.now14}.
Among many application areas for deep learning, the most prominent ones are agricultural inspection for fruit counting \cite{sa_mccool.sensors16} and disease detection \cite{mohanty_salathe.fps16}, vehicle and pedestrian traffic monitoring \cite{sermanet_lecun.cvpr13, lv_wang.its15}, structural health monitoring of critical infrastructure \cite{cha_buyukozturk.cacie17, yeum2015vision, makantasis2015deep} to name a few.

MAVs are versatile platforms for collecting imagery which can be used to train and test deep networks.
These platforms are especially cut out for applications which require multiple traversals of large volumes with hard-to-reach spots for a human worker such as large infrastructure like tunnels, bridges and towers \cite{ozaslan_kumar.ral18, quenzel2018autonomous, burri2012aerial}.
Collecting large image datasets required for training a deep network from such environments has become possible with the emerge of complete estimation and navigation stacks for MAVs \cite{ozaslan2017autonomous, mohta2018fast, loianno2017estimation}.
This work relies on such datasets collected by \cite{ozaslan2017autonomous}.

Critical infrastructure such as dams, bridges and skyscrapers experience structural deterioration due to corrosion, aging and tremendous repetitive loads.
Other external factors exacerbating this problem are earthquakes and adverse extreme atmospheric conditions such as storms.
In order to avoid possible catastrophic consequences such as demolishment of these infrastructure, flood and fire, periodical inspection and maintenance are indispensable.
The situation is far more severe for dams and penstocks since many hydraulic power plants and water conduits in The U.S. were built more than a half century ago \footnote[3]{https://www.infrastructurereportcard.org/cat-item/dams/}.

Penstocks are steel or concrete tunnels that carry water from upstream to turbines at the bottom of a dam to generate electricity.
Visual inspection of a penstock is possible only when it is completely dewatered which in turn interrupts electric generation and downstream regulation.
Conventional inspection methods require building a scaffolding inside the penstock and an inspector to either climb up inside the penstock or swing down from the gate to spot regions on the tunnel surface that require maintenance using only a hand-held torch.
However, these methods are time-consuming, labor-intensive, dangerous, inaccurate due to difficult low-light working conditions and rely on the inspector's subjective decision.
For these reasons, it is crucial to perform visual inspection faster and more accurate with the least human intervention to reduce maintenance cost, safety threads and increase effectiveness.

\begin{figure}[t]
	\centering
	\includegraphics[width=\linewidth]{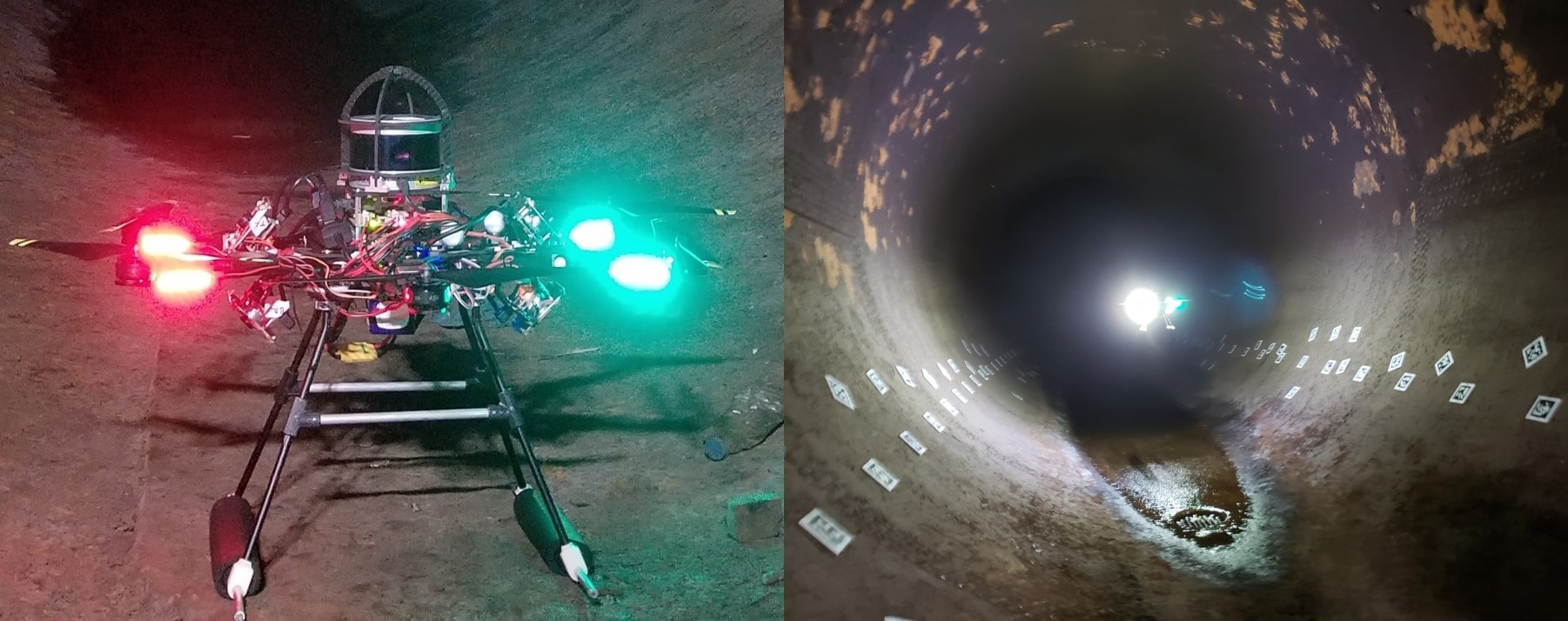}
	\caption{
		The MAV platform flying inside a penstock at Center Hill Dam, TN. 
		The robot collects images from four onboard cameras with the only illumination source being the onboard power LEDs.
	}
	\label{fig:drone}
	\vspace{-3mm}
\end{figure}

In this study, we propose a data-driven, deep learning method to automatically identify corroded spots on the surface of a penstock.
The proposed U-Net design requires a small training dataset consisting of less than 40 manually annotated image samples.
This network performs pixel-wise classification of image regions into groups such as coating, water, rivet, wet or corroded surface.
The off-line classification algorithm runs at 12 frame-per-second (fps) on the original $1024 \times 1280$ images which offers real-time, onsite rapid processing of high-resolution raw images.
Fig. \ref{fig:intro_image} shows an example output of our classifier. 

We rely on the datasets collected by \cite{ozaslan2017autonomous} which uses an autonomous MAV to traverse a penstock while collecting imagery from four onboard color cameras effective field of view of which covers the whole annulus of the tunnel.
In this respect, the proposed method complements \cite{ozaslan2017autonomous} in providing an end-to-end autonomous tunnel inspection system.
Our U-Net works successfully despite low-light conditions and excessive noise due to dust and mist particles occluding camera views (Fig. \ref{fig:qualitative_result}).
To our knowledge, this is the first study that can do automated defect detection on a dataset collected autonomously from such critical infrastructure using an MAV.

In the subsequent sections, we discuss the class imbalance problem and consider different loss functions to mitigate this problem. 
We generalize the focal loss function which is originally proposed for a single class problem \cite{lin2018focal} for our multi-class segmentation problem.
Also, we analyze our experimental results with different metrics to signify the importance of choosing the right metrics in performance evaluation of the learned models.
Our empirical experiments show that the focal loss function, associated with a proper set of class weights can improve the segmentation results.
We also discuss the effect of the class weight on the performance of the weighted focal loss. 

This work presents multiple contributions :
\textit{First}, we extend the focal loss function to handle multi-class segmentation problem which is originally proposed for single class detection. 
\textit{Secondly}, the proposed U-Net is retrofitted to work in low-light conditions and in presence of excessive image noise due to dust and mist particles causing significant camera occlusion. 
\textit{Finally}, this work complements our previous work \cite{ozaslan2017autonomous} in offering an end-to-end autonomous tunnel inspection system consisting of autonomous data collection with an MAV and automated image annotation with minimal input from human operators for both flying the MAV and training the deep network.
To our knowledge, this is the first study that offers a complete solution to inspection of such critical large infrastructure under challenging low-light and high-noise environments.




%% file: tex/related_work_tolga.tex
\section{Related Work}
\label{sec:related_work}

\subsection{Deep Learning for Visual Infrastructure Inspection}


There have been a huge interest in using deep learning techniques for infrastructure inspection.
In one of the recent studies, \cite{cha_buyukozturk.cacie17} introduces a sliding-window technique using a CNN-based classifier to detect cracks on concrete and steel surfaces. 
The major drawback of this method is that it cannot satisfy real-time processing requirements and would fail in detecting small defects which are very frequent in our images. 
This type of framework is also not data-efficient since it proceeds an image patch as a single sample. 
Indeed, authors mention that they train with 40K image patches cropped out from 277 high resolution images captured with a hand-held camera which is far beyond the number of training images we use in this work.

\cite{park2016machine} uses CNNs to detect defects on different types of materials such as fabric, stone and wood.
They compare their network to conventional machine learning methods such as Gabor Filter, Random Forest and Independent Component Analysis.
However, the authors do not aim at all for a complete system for autonomous data collection and automated defect detection.
Finally, their training dataset consists of 3000 images which is also far larger than ours.

In a similar application to ours, \cite{makantasis2015deep} proposes to feed a CNN with low-level features such as edges so as to obtain a mixture of low and high level features before classifying pixels to detect defects on concrete tunnel surfaces. 
Unlike this work, we propose an end-to-end fully convolutional neural network that does not depend on handcrafted features which also works with arbitrary input image sizes. 
Indeed, some of the low-level features used in their work are neither easy to obtain nor provide useful information for the CNN such as edges, texture and frequency.
This is due to the characteristics of the noise caused by dust and mist as well as the complex appearance of the penstock surface.




\subsection{Small training dataset}

The quality and size of training data is crucial for supervised learning tasks such as image segmentation.
The data size gains more importance for deep neural networks which are known to require big train sets. 
However, amount of training data is limited in field robotics applications due to the cost of the data collection process. 
To mitigate this problem, several solutions have been proposed such as generating synthetic data \cite{georgakis2017synthesizing}, data augmentation \cite{chen2018deeplab}, using transfer learning \cite{romera2018train} and designing data-efficient approaches, to name a few. 

A well-known successful data-efficient deep neural networks is U-Net \cite{ronneberger2015u} which won the Cell Tracking Challenge at ISBI 2015 \footnote{http://www.celltrackingchallenge.net/}.
U-Net can be trained using only as few as $30$ data samples. 
Due to the above considerations, in this work, we adopt this network design after some simplifications.

Another work related to our segmentation task is \cite{guerrero2018multiclass} on biomedical imaging. 
The authors focus on instance segmentation which attempts to label individual biological cells. 
Their problem is also less challenging, due to well-controlled lighting conditions which is not the case in our application as seen in Fig.~\ref{fig:intro_image}.

\subsection{Class Imbalance in Deep Learning}

Class imbalance, on which exists vast literature in classical machine learning domain, has not attracted significant attention in the deep learning context.
One of the very few studies on this topic is by Buda et. al \cite{buda2018systematic} who provide a systematic investigation on class imbalance in deep learning.
However, their solutions focus on redistributing the frequency of classes in the training dataset using sampling, two-phase training and thresholding rather than loss functions like us. 

Authors of \cite{lin2018focal} propose a focal loss as an alternative to sampling methods. 
However, it is meant to solve the single-class detection problem. 
In this study, we extend the horizon of its potential usage by investigating its use in a multi-class segmentation problem. 

Unlike the above studies, our images suffer from imperfect lighting conditions such as high exposure on reflective areas, low exposure under non-reflective areas.
Furthermore, the propeller downwash kicks dust and mist which occludes the camera view.
In addition, the corroded spots are highly nonhomogeneous in appearance and size, making the segmentation further challenging. 

%% file: tex/dataset.tex
\section{Dataset}
\label{sec:dataset}

\begin{figure*}[t]
  \vspace{3mm}
    \begin{minipage}{.50\textwidth}
      \raggedleft  
      \subcaption{}
      \vspace{-2mm}
      \includegraphics[width=\linewidth]{{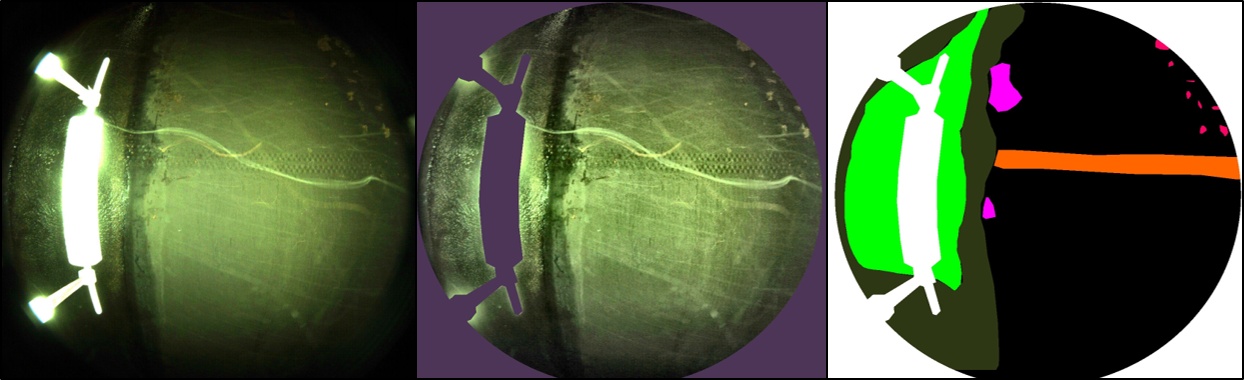}}
    \end{minipage}
    \hfill
     \begin{minipage}{.48\textwidth}
     \raggedright
     \subcaption{}
      \vspace{-2mm}
      \includegraphics[width=\linewidth]{{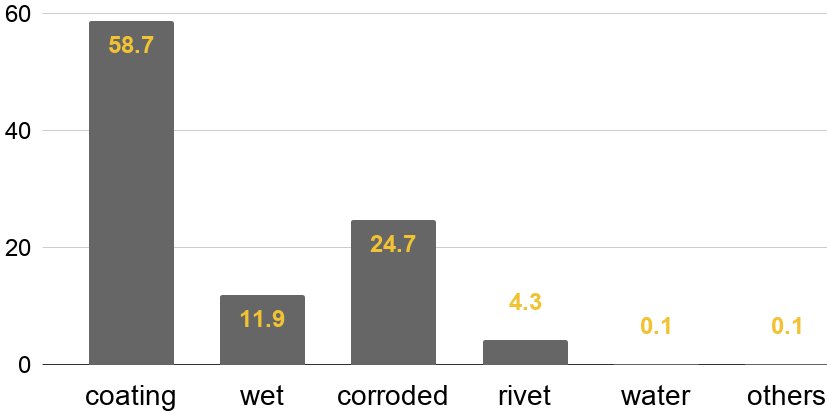}}
    \end{minipage}  \\
    \caption{(a): Sample of training image. From left to right: raw image, image after histogram equalization and masking, labeled image. Red: corroded, Pink: corroded (noisy labelling), orange: rivet, green: water, dark green: wet region, black: normal coating; some light traces occur due to flowing dust. (b): Class distribution of pixels in percentage.}
    \label{fig:intro_image}
    \vspace{-3mm}
\end{figure*}

\subsection{Data Collection}

The dataset used in this study is collected with a customized DJI-F550 platform described in \cite{ozaslan2017autonomous} that autonomously flies inside a penstock at Center Hill Dam, TN. 
The MAV, shown in Fig.~\ref{fig:drone}, is equipped with four synchronized color cameras mounted such that the combined field of view covers the annulus of the tunnel.
A typical dataset contains 3600 color images from each camera saved in a compressed format. 

\subsection{Data Preprocessing}

Weak on-board illumination, reflective wet surfaces and noise due to dust and mist particles cause images to be either pale and textureless, very bright or occluded with bright dust traces (Fig. \ref{fig:qualitative_result}).
Hence, we preprocess the images to suppress these effects before feeding them into our network.
We apply limited adaptive histogram equalization using OpenCV's CLAHE module to mitigate the brightness imbalance. 
Also, image regions occluded by MAV's landing gear and camera lens covers are masked out.
In order to do inference in the first-person view, we omit image undistortion.

We pick 70 images, 38 of which is used for training. The training images are extracted from the first portion of the video while the test images are extracted from the later. 
In order to make sure that the images are captured from different view points, we pick one every $40$ frames, starting from the moment the robot reaches its horizontal velocity of $\sim2~m/s$.
Fig. \ref{fig:intro_image}a shows a sample labeled image. 

We label pixels in an image according to $6$ different classes including: normal coating, wet coating, corroded, rivet, water and others.
Fig. \ref{fig:intro_image}b shows the percentage of these classes in the training dataset. 

At this point we have to emphasize that although the total corroded spots constitute $24.7 \%$ of the unoccluded pixels, the human data labeler was confident only for $7\%$ of the whole set. 
For $17.7 \%$ of it, the labelers was not confident with their choice due to lack of image details.
The latter group of image regions are marked by pink color in Fig~\ref{fig:intro_image}a. 
One can ignore the pixels belonging to these regions in the training set to avoid noisy labeling at the expense of loosing positive samples for training. 
In this study, we are aware that these regions can cause problems and use them with caution as corroded spots.

%% file: tex/problem.tex
\section{Problem Definition}
\label{sec:problem}

\subsection{The Multi-class Segmentation Problem}
Our goal is to detect corroded spots from images captured by an autonomous MAV flying inside a penstock.
There are various object detection and semantic segmentation techniques for single-class or multi-class segmentation that could possibly work for this purpose. 
However, we chose semantic segmentation rather than object detection because corroded spots in our data set are highly variant in size, shape, position, and intensity value, making it challenging to detect them using bounding boxes. 

We formulate this problem as a multi-class segmentation task.
Let $\mathbb{I}_{train} = \left\{ \mathbf{I}_1, \mathbf{I}_2, \dots \mathbf{I}_N \right \}$ be the set of training images which have the dimension of $h \times w \times d$. 
Each image $\mathbf{I}_i$ is associated with a mask $\mathbf{M}_i$ of size $h \times w \times 1$ that tells the class every pixel on the image $\mathbf{I}_i$ belongs to, assuming that each pixel belongs to only a single class.  
More specifically, a pixel $x_{ij}, i = 1\dots h, j=1\dots w$ on an image $\mathbf{I}_n, n=1\dots N$ has an intensity value of $\mathbf{I}_n(x_{ij}) \in \mathbb{R}^d$ and a label $\mathbf{M}_n(x_{ij}) = 1\dots \mathbf{C}$, where $\mathbf{C}$ is the number of classes considered in the problem. 

During the training process, the network is fed with samples which are pairs $<\mathbf{I}_n(x_{ij}),  \mathbf{M}_n(x_{ij})>$. 
Essentially, the network attempts to map $\mathbf{I}_n(x_{ij})$ to a label $ \mathbf{M}_n(x_{ij})$ but it is only able to achieve an estimate $\hat{\mathbf{M}}_n(x_{ij})$. 
The difference between $\hat{\mathbf{M}}_n(x_{ij})$ and $\mathbf{M}_n(x_{ij})$ indicates how good the current model is and provides a training signal to adjust the model's hyperparameters accordingly.  
During testing, the network is only fed with samples $\mathbf{I}_n(x_{ij}) \in \mathbb{I}_{test}$ and attempts to predict the corresponding labels. 


\subsection{Loss Function and Updating Parameters of Deep Network}

In a supervised deep learning framework, a loss function acts as a measure of the goodness of the current model state during the training process. 
Let $\textbf{y}$ and $\hat{\textbf{y}}$ be the ground-truth labels and the network outputs.
The training process can be considered as a highly non-linear optimization problem that minimizes a loss or objective function
$\mathbf{L}( \textbf{y}, \hat{\textbf{y}})$. 
We do this by iteratively updating the network parameters, $\mathbf{W}$, as a function of the derivative of the loss function with respect to $\mathbf{W}$ which is written by the recurrence relation
\begin{equation}
    \mathbf{W}_{i+1} = \mathbf{W}_i + \eta \frac{d \mathbf{L}}{d \mathbf{W}}.
    \label{eq:update}
\end{equation}
where $\mathbf{W}_i$ represents the parameters of the network at iteration $i$ and $\eta$ is the learning rate.
Thus, the choice of the loss function is one of the determining factors on how a deep network frame performs since with the right loss function the training process can converge much faster and also result in a network that can do more accurate inference on the test data.

%% file: tex/class_imbalance.tex
\section{Class Imbalance and Loss Functions}
\label{sec:class_imbalance}
 
\subsection{Class Imbalance}

Each data sample in the training dataset contributes to the update Eq.~\ref{eq:update} through the loss function, $\mathbf{L}$, regardless of which training scheme is used during training such as stochastic gradient descent and mini-batch stochastic gradient descent. 
Thus, the frequency of a class in the training dataset determines the shape of a given loss function, i.e. the more samples a class contains, the more the class affects the loss function which therefore affects the training process. 
A dominant class in the training dataset contributes much more to the loss function, $\mathbf{L}$, than other class samples.
This then biases the training process in a way that the network assigns all training data to the dominant class for the sake of minimizing the loss function.
Consequently, the trained model could wrongly predict all test samples to be belonging to this class.
This problem is called \textit{class imbalance}. 

Generally speaking, class imbalance is a common problem in tasks such as object detection and segmentation. 
In robotics and medical image applications, this problem is extremely critical since the training data set is often small due to the expensive data acquisition process.



To mitigate this problem, there have been extensive works on strategic sampling. 
In this study, we generalize the focal loss, introduced in \cite{lin2018focal}  as an alternative to these sampling-based methods, to multi-class segmentation problems.


\subsection{Softmax Cross Entropy (SCE) and Weighted Softmax Cross Entropy (W-SCE)}

Let $\mathbf{C}$ be the number of classes considered in the classification problem and $\mathbf{N}$ be the number of samples used to calculate the loss to update the network parameters. 
Let $y_{n}$, $\hat{y}_{n}$ with $n=1 \dots \mathbf{N}$ be the one-hot vector of the true labels and the corresponding softmax output from the network respectively. 
The softmax cross entropy loss can be defined as 
\begin{equation}
\mathbf{L}_{SCE} = - \frac{1}{\mathbf{N}} \sum_{n=1}^\mathbf{N} \sum_{c=1}^\mathbf{C} y_{nc} \log{\hat{y}_{nc}}
\end{equation}
where $\hat{y}_{nc}$ is essentially the network's confidence of the sample $n $ being classified as class $c$ and $\sum_{c=1}^\mathbf{C} y_{nc} = 1$. 

To address the class imbalance problem, one common trick is to associate weighting factors $w_c \in [0, 1]$ with $c=1 \dots \mathbf{C}$ for classes. 
These $w_c$ can be set by inverting the class frequency in the training data set or by tuning as hyperparameters. 
In this study, we use the former method. 

The cross entropy loss becomes weighted cross entropy loss
\begin{equation}
\mathbf{L}_{W-SCE} = - \frac{1}{\mathbf{N}} \sum_{n=1}^\mathbf{N} \sum_{c=1}^\mathbf{C} w_c y_{nc} \log{\hat{y}_{nc}}.
\label{eq:L_BCE}
\end{equation}
An advantage of this loss function is that it can emphasize the importance of rare classes or class of concern  in the loss function, providing better training signals to the network. 
However, this loss function cannot differentiate between easy and hard samples. 

\subsection{Focal Loss for Multi-class Classification}

We can generalize focal loss \cite{lin2018focal} for multi-class classification problem by replacing $w_{c}$ factor in Eq.~\ref{eq:L_BCE} with a weighting factor $(1-\hat{y}_{nc})^\gamma$, where $\gamma > 0$ is a tunable parameter which writes as
\begin{equation}
\mathbf{L}_{Focal} = - \frac{1}{\mathbf{N}} \sum_{n=1}^\mathbf{N} \sum_{c=1}^\mathbf{C} (1-\hat{y}_{nc})^\gamma y_{nc} \log{\hat{y}_{nc}}
\label{eq:L_focal}
\end{equation}

The effect of the focal loss and $\gamma$ value can be explained as follows:
When a hard sample is misclassified with low confidence on the true class, i.e. $\hat{y}_{nc}$ is small, the weighting factor becomes close to $1$ preserving that sample's contributions to the total loss. 
In contrast, an easy sample correctly classified with a high confidence value, $\hat{y}_{nc}$, will have its weight close to $0$ reducing its contribution to the total loss. 
In summary, the focal loss function can appreciate the importance of hard samples, regardless of which class they belong to, by giving more them more weight and downweight  easy samples. 

According to the authors, the focusing parameter $\gamma$ controls the rate at which how easy samples are downweighted over time. 
In the special case of $\gamma = 0$, the focal loss becomes equivalent to cross entropy.
When $\gamma$ is high, the weighting factor is exponentially small, extending the range of samples considered as easy samples. 

\subsection{Weighted Focal Loss (W-Focal)}

A drawback of focal loss function is that it can underestimate the importance of samples in the classes of concern. 
In addition, it is sensitive to wrong labeled samples in the training dataset since the wrong-labeled samples would mistakenly be considered as hard samples. 
We discuss these problem in more details in Sec.~\ref{sec:evaluation}. 



Thus, in addition to $\mathbf{L}_{Focal}$, we also investigate the performance of weighted focal loss which writes as 
\begin{equation}
\mathbf{L}_{W-Focal} = - \frac{1}{\mathbf{N}} \sum_{n=1}^\mathbf{N} \sum_{c=1}^\mathbf{C} w_c (1-\hat{y}_{nc})^\gamma y_{nc} \log{\hat{y}_{nc}}. 
\label{eq:L_alpha_focal}
\end{equation}
Essentially, $\mathbf{L}_{W-Focal}$ can solve the aforementioned problem that focal loss suffers from by adjusting $w_c$ to emphasize the importance of a certain class $c$, as well as reducing $w_c$ associated with classes that might have been labeled wrongly.

%% file: tex/training.tex
\section{Training}
\label{sec:training}
\subsection{U-Net}

\begin{figure*}[ht]
\centering
\includegraphics[width=0.7\linewidth]{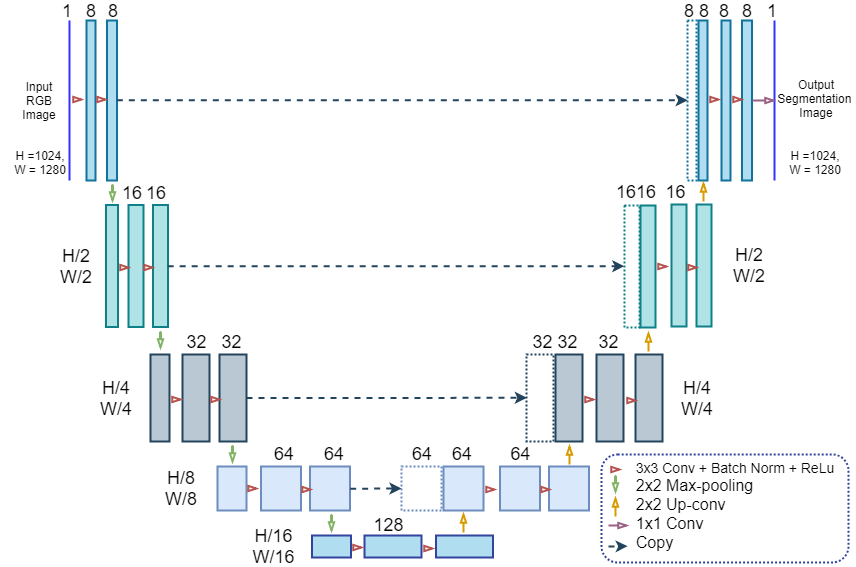}
\caption{Architecture of the designed U-net}
\label{fig:unet}
\end{figure*}

There have been successive Fully Convolutional Networks (FCNs) for image Semantic Segmentation followed by~\cite{long2015fully} such as U-Net~\cite{ronneberger2015u},  Deepnet~\cite{chen2018deeplab}, and Segnet~\cite{badrinarayanan2015segnet}. 
In this study, we adopt U-Net~\cite{ronneberger2015u} as it has shown superior performance in biomedical applications.
Also, since U-Net design is simple, we could focus more on investigating on alternative loss functions and their performance with small training dataset and class imbalance.




Unlike the original U-Net, we reduce the number of features on each block by $8$ on each block such that the inference during testing can be done in real time. 
We also make use of batch-norm \cite{ioffe2015batch} on every convolution layer as a means to achieve better model regularization.  

\subsection{Training Scheme}

The deep network is implemented in Tensorflow \cite{abadi2016tensorflow} using mini-batch gradient descent with the batch-size of 2. 
We use an Adam Optimizer with default parameters. 

To cope with the small training dataset, we intensively utilize data augmentation techniques including random rotation, random cropping and padding, random gamma shifting, random brightness shifting, and random color shifting. 

In order to evaluate the performance of every loss function, we create multiple variances to train and test.
For each variance of the model, we associate U-Net with one of the following loss functions: 
\begin{enumerate}
    \item Softmax cross-entropy (SCE) 
    \item Weighted softmax cross-entropy (W-SCE) 
    \item Focal loss (Focal). We evaluate three variances correponding to $\gamma = 0.5, 1.0, 2.0$. 
    \item Weighted focal loss (W-Focal). We evaluate three variances correponding to $\gamma = 0.5, 1.0, 2.0$. 
\end{enumerate}
While the loss functions in the variances are different, other settings are kept the same. 
The weights $w_c$ for pixel classes are chosen to be approximately inverse of their frequencies in the training dataset, except the for classes $water$ and $others$ since their annotation is too noisy. 
Indeed, our set of weights is 
$coating: 1$, $wet coating: 1$, $corroded: 10$, $rivet: 5$, $water: 1$, $others: 1$.

Each variance is then trained separately from scratch over $1800$ epochs with early stop executed based on Dice similarity coefficient value (Sec.~\ref{sec:evaluation}) on the evaluation dataset.

%% file: tex/evaluation.tex
\section{Results and Evaluation}

\label{sec:evaluation}

In this section, we first discuss the metrics that we use in evaluation. 
Then, we present our experimental results and discuss the performance of different loss functions from both qualitative and quantitative perspectives. 
Since corrosion detection is the primary concern of this work, we present all metrics and evaluations for the \textit{corroded} class.

\subsection{Evaluation Metrics}
Choosing a proper evaluation metric for a segmentation task is of big importance since this will decide which models are favored. 
According to Csurka et. al~\cite{csurka2013good}, judging a segmentation algorithm is difficult since it is highly application dependent.





In this study, our quantitative results are reported in terms of four metrics: Dice similarity coefficient (DSC),  sensitivity, specificity, and total error which write as
\begin{align}
  \mathbf{DSC} &= \frac{2TP}{2TP + FP + FN} \\ 
  \mathbf{Sensitivity} &= \frac{TP}{TP + FN} \\
  \mathbf{Specificity} &= \frac{TN}{TN + FP} \\
  \mathbf{Total} \mathbf{Error} &= \frac{\alpha}{\alpha + 1}   FN + \frac{1}{\alpha + 1} FP
\end{align}
where, $TP=$ Pixels correctly classified as corroded in the ground truth and by algorithm; $FP=$ Pixels not classified as corroded in the ground truth, but classified as corroded by algorithm; $TN=$ Pixels not classified as corroded in ground truth and by algorithm; $FN=$ Pixels classified as corroded in ground truth, but not classified as corroded by algorithm.

While \textit{sensitivity} is a measure of $TP$ rate, \textit{specificity} is a measure of $TN$ rate.
\textit{DSC} can reflect the $TP$ rate as well as penalize $FP$ and $FN$. 
On the other hand, \textit{total error} introducing an adjustable $\alpha$ value, can provide a flexible measure to reflect how much more missing a $TP$ costs than missing a $TN$. 
Note that total error does not take into account the $TP$ as DSC does and total error is dependent on the $\gamma$.
Thus, DSC and total error do not necessarily agree on judging which model is better.

In our experiments, $\alpha$ is set to be $10$, addressing the importance of identifying corroded spots.
We choose the favorable model as the one with high DSC and acceptable total error.

\subsection{Quantitative and Qualitative Results}
\begin{table}[t]
\centering
\begin{tabular}{|c|c|c|c|c|c|}
\hline
Loss    & $\gamma$ & DSC  & Sensitivity  & Specificity & Total Error  \\ \hline
\multirow{3}{*}{Focal} 
& $2.0$ & $52.5$  & $67.9$  & $98.0$ & $2.9$ \\ \cline{2-6}
& $1.0$ & $51.1$ & $57.5$ & $97.8$ & $3.6 $ \\ \cline{2-6} 
& $0.5$ & $49.6$  & $67.5$  & $97.0$ & $3.2$\\ \hline
\multirow{3}{*}{} 
  W- & $\mathbf{2.0}$ & $\mathbf{52.1}$ & $\mathbf{73.6}$  & $\mathbf{97.7} $ & $\mathbf{2.2}$ \\ \cline{2-6}  
 Focal & $1.0$ & $49.5$  &  $75.2$   &  $97.2$ & $2.1 $  \\ \cline{2-6} 
    & $0.5$ & $ 49.5$  & $78.7$ & $96.4$ & $1.9$ \\  \hline
\multicolumn{2}{|c|}{SCE} 
&$50.6$ & $67.5$ & $97.3$ & $2.9$ \\ \hline
\multicolumn{2}{|c|}{W-SCE}     
& $51.1$ & $75.2$ & $97.0$ & $2.5$  \\ \hline
\end{tabular}
\caption{Performance with fixed class weights ($\%$)\\
DSC, Sensitivity, Specificity: Higher is better. Total Error: Smaller is better}
\label{tab:quant_result}
\vspace{-3mm}
\end{table}

\begin{figure*}[t!]
  \vspace{5mm}
  \centering
    \begin{minipage}{.24\textwidth}
      \vspace{-2mm}
      \includegraphics[width=\linewidth]{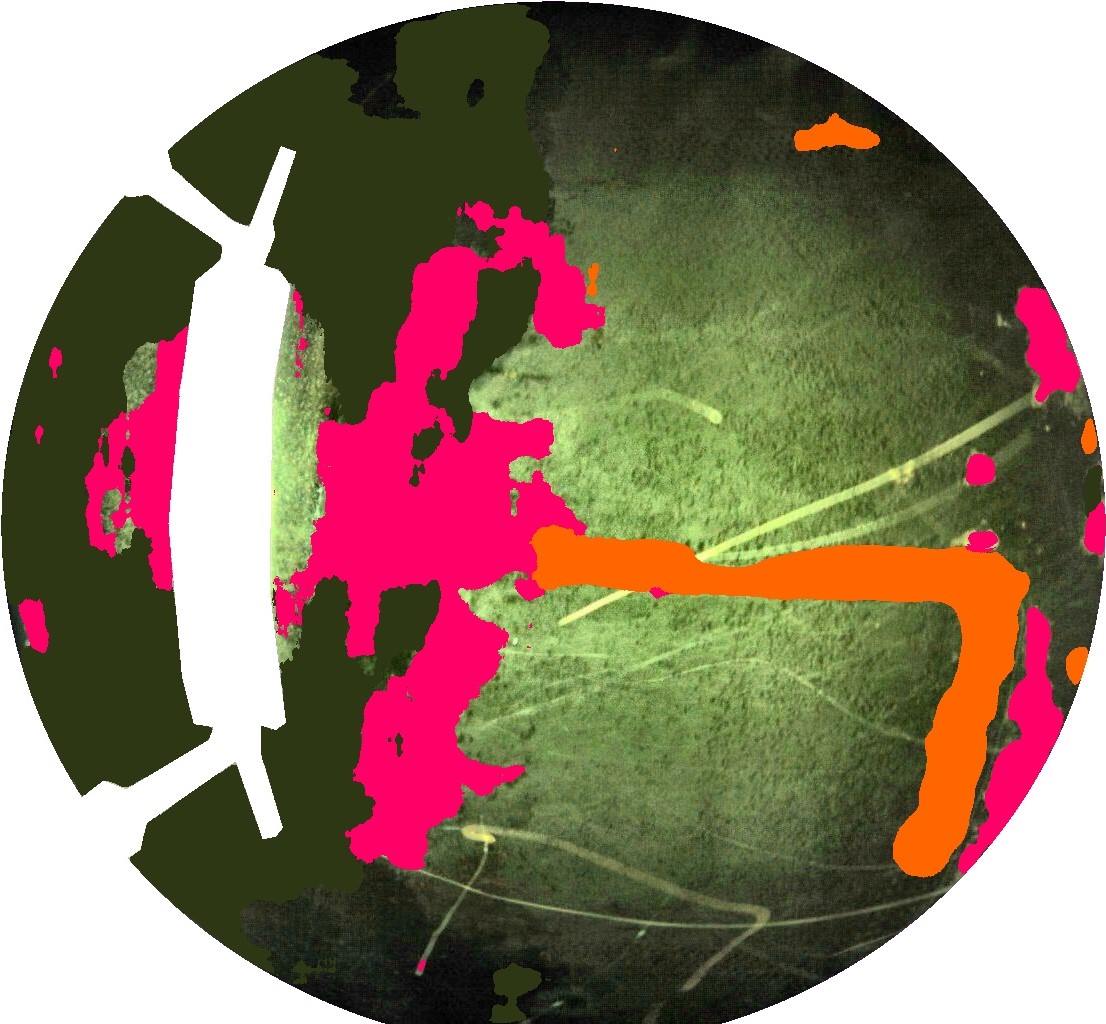}
    \end{minipage}
    \begin{minipage}{.24\textwidth}
      \vspace{-2mm}
      \includegraphics[width=\linewidth]{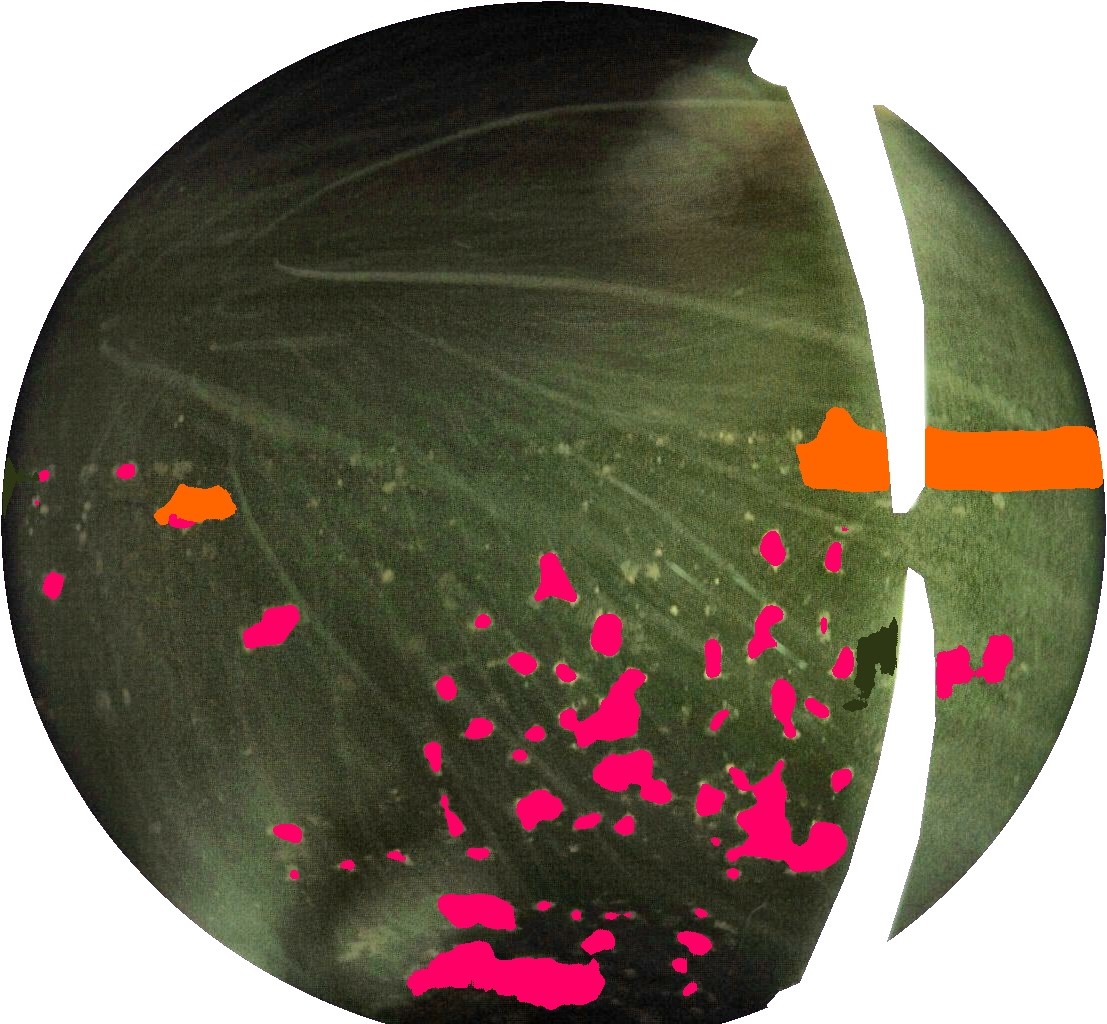}
    \end{minipage}
    \begin{minipage}{.24\textwidth}
      \vspace{-2mm}
      \includegraphics[width=\linewidth]{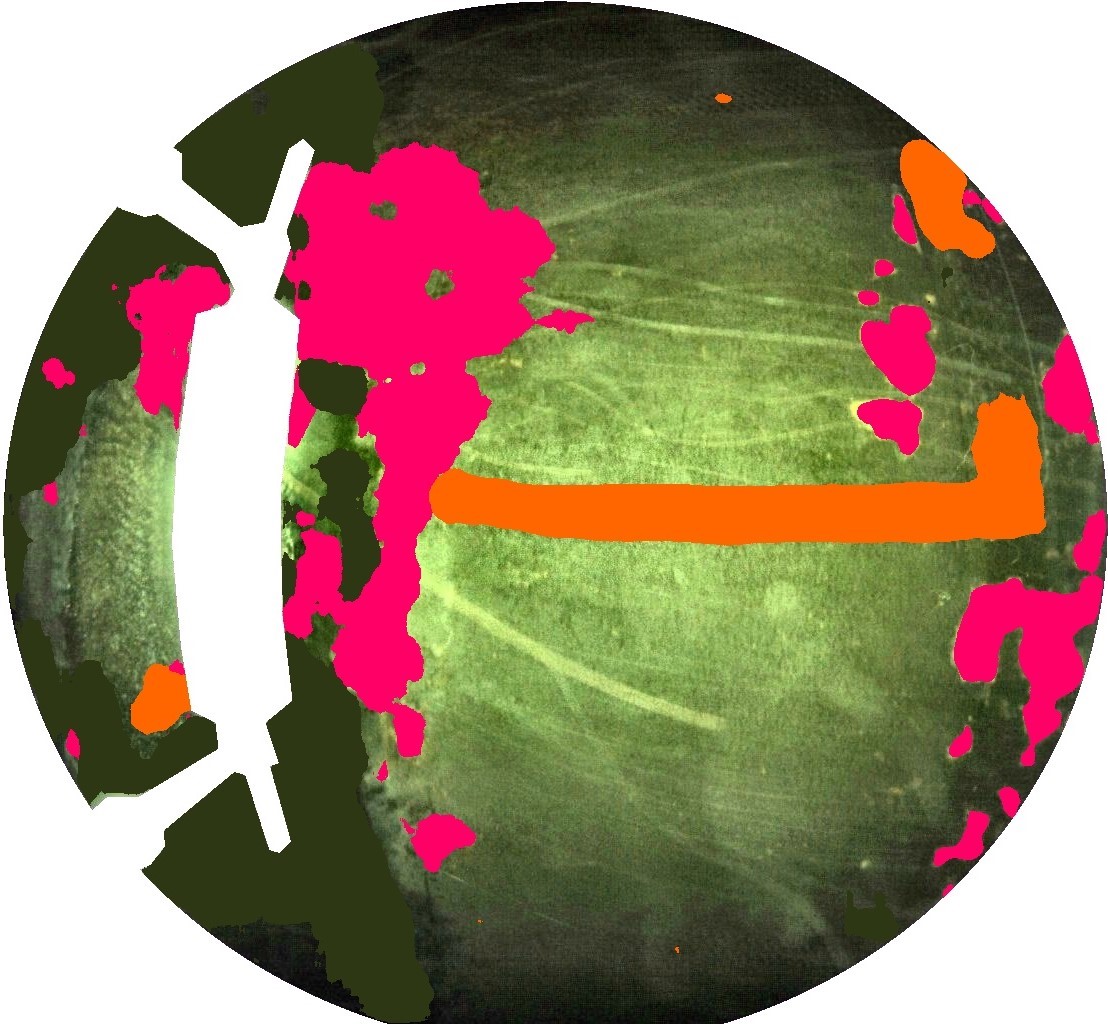}
    \end{minipage}
     \begin{minipage}{.24\textwidth}
      \vspace{-2mm}
      \includegraphics[width=\linewidth]{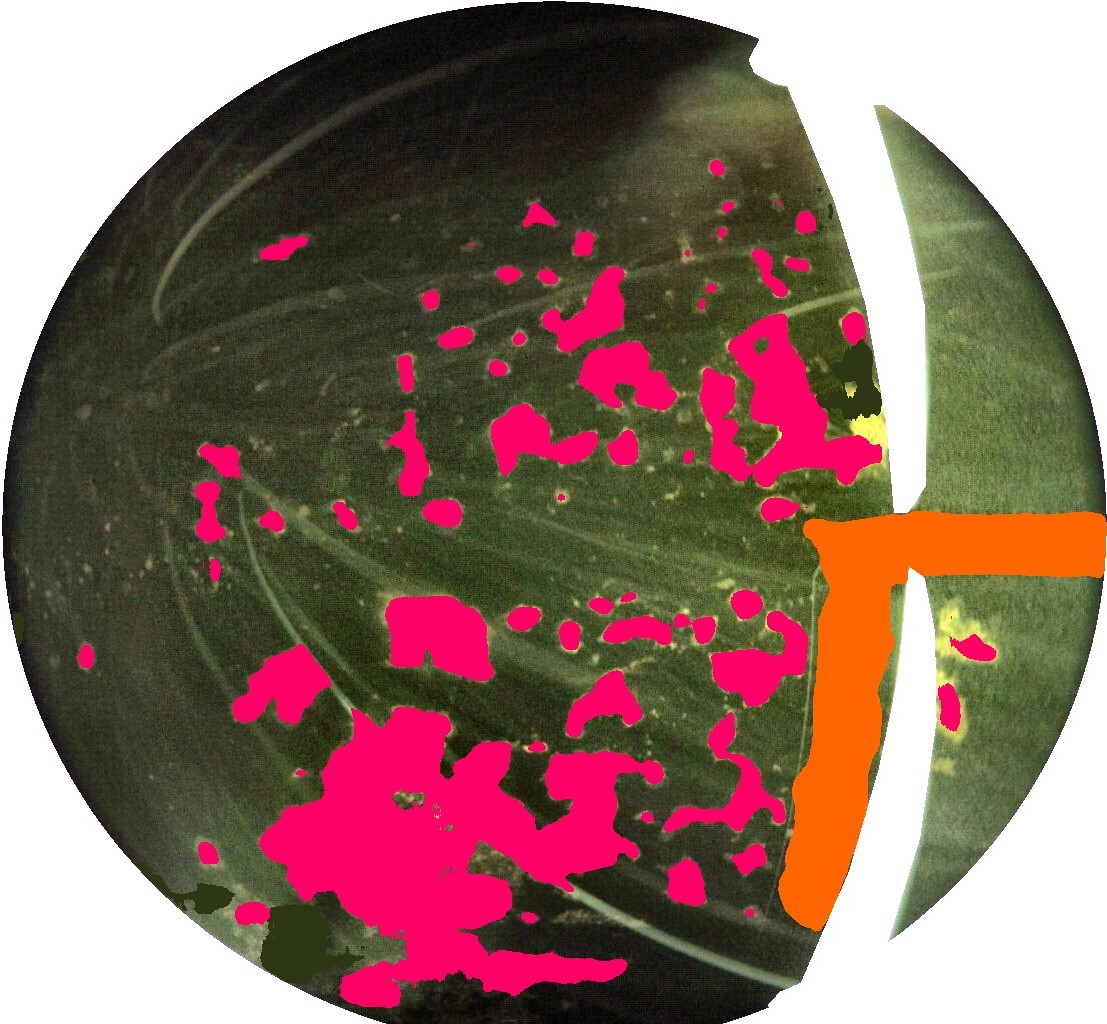}
    \end{minipage}  \\
\vspace{.1in}
    \begin{minipage}{.24\textwidth}
      \vspace{-2mm}
      \includegraphics[width=\linewidth]{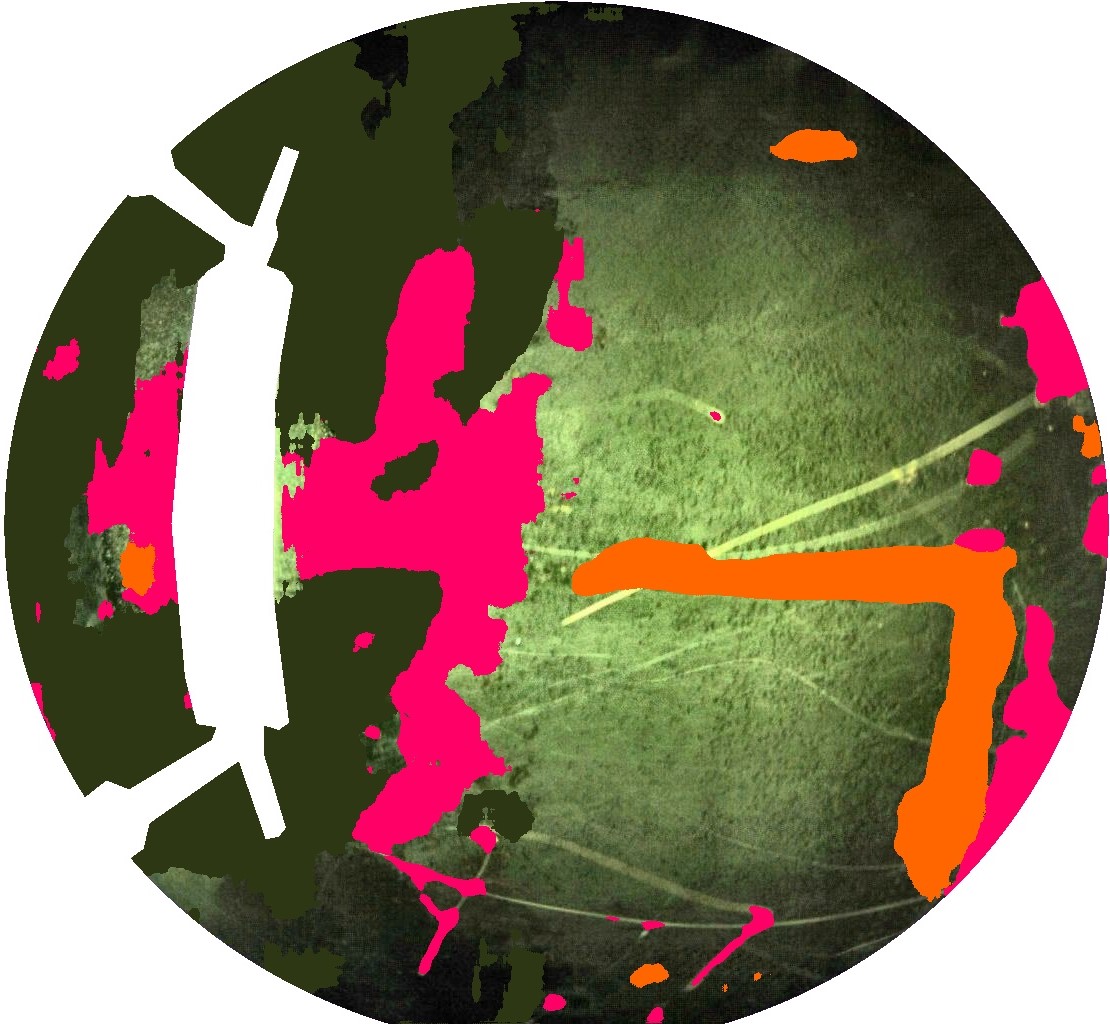}
    \end{minipage}
    \begin{minipage}{.24\textwidth}
      \vspace{-2mm}
      \includegraphics[width=\linewidth]{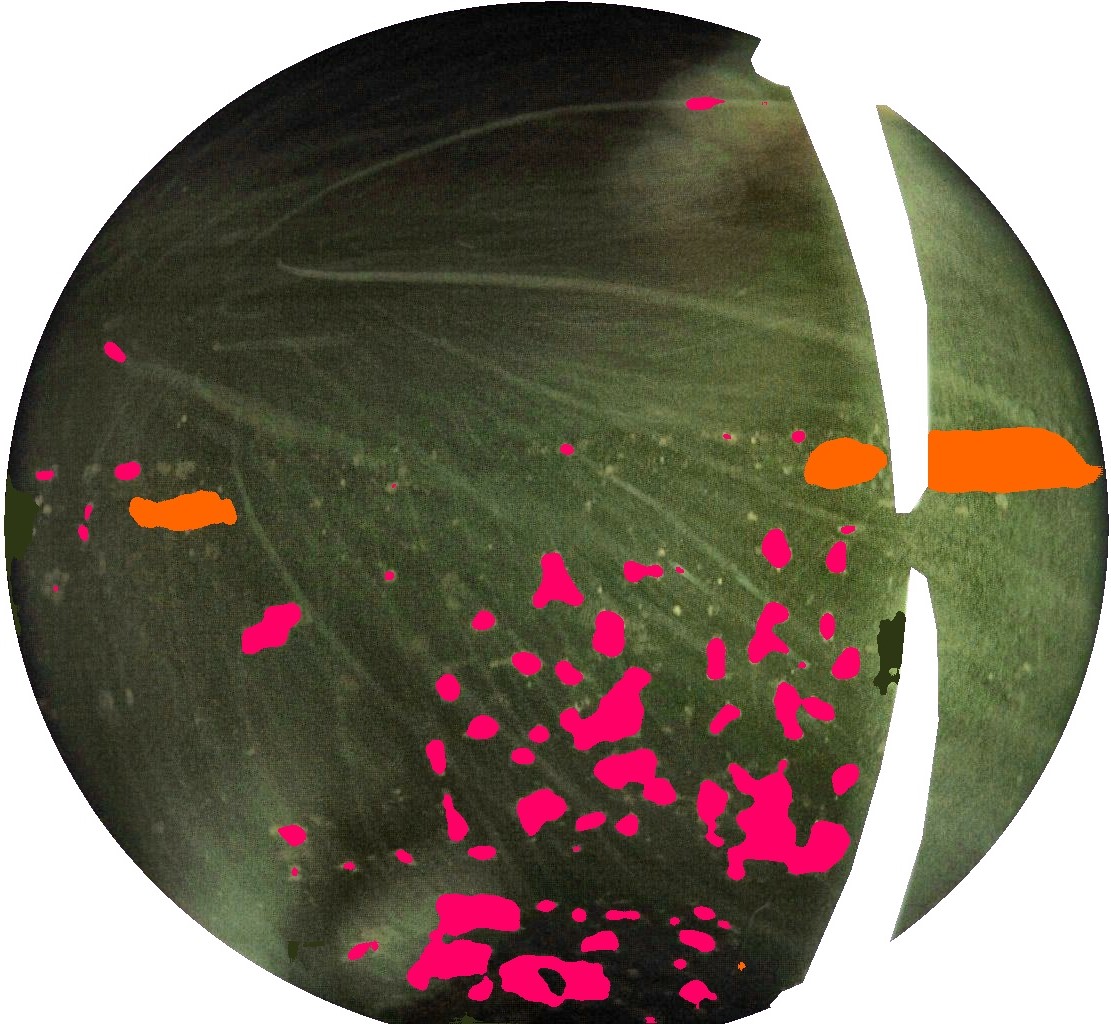}
    \end{minipage}
    \begin{minipage}{.24\textwidth}
      \vspace{-2mm}
      \includegraphics[width=\linewidth]{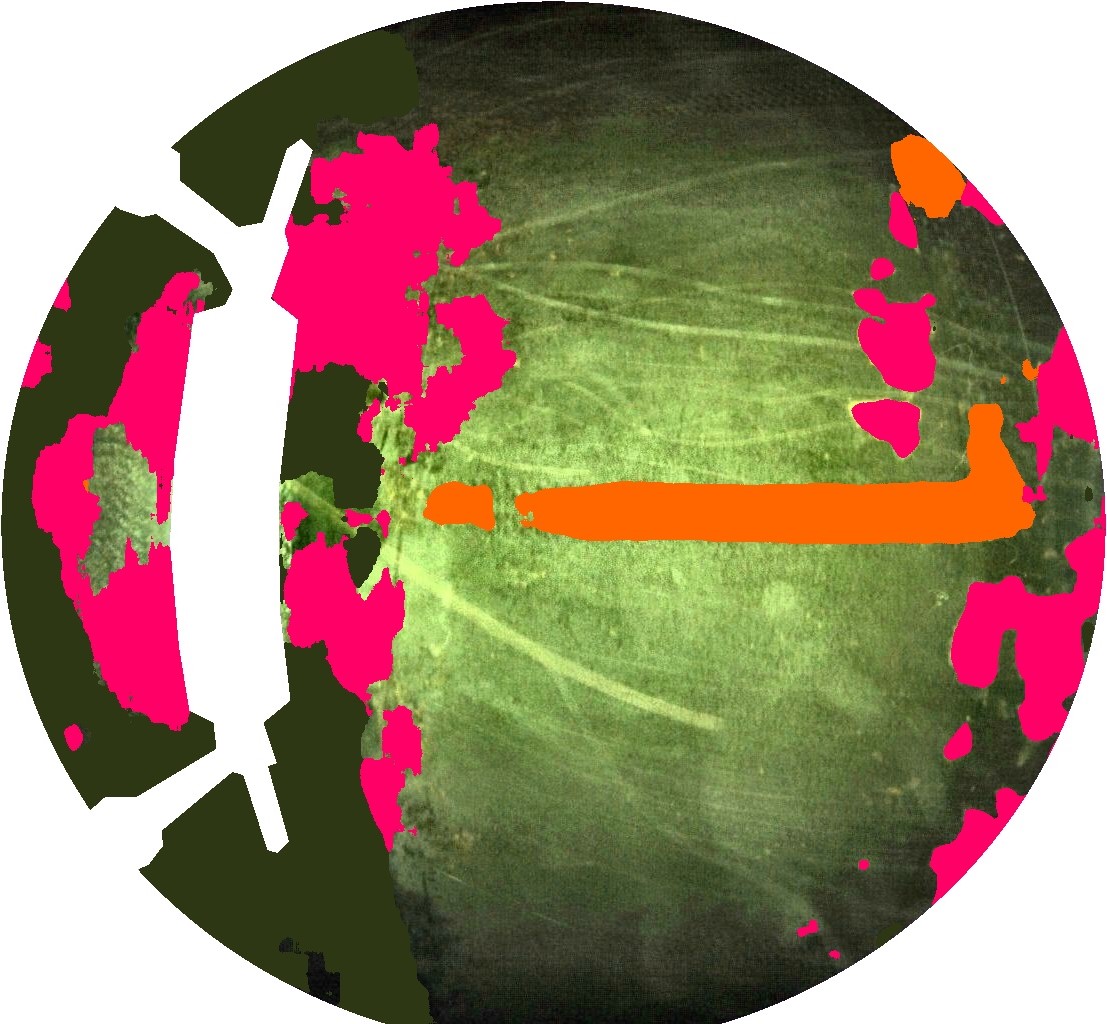}
    \end{minipage}
     \begin{minipage}{.24\textwidth}
      \vspace{-2mm}
      \includegraphics[width=\linewidth]{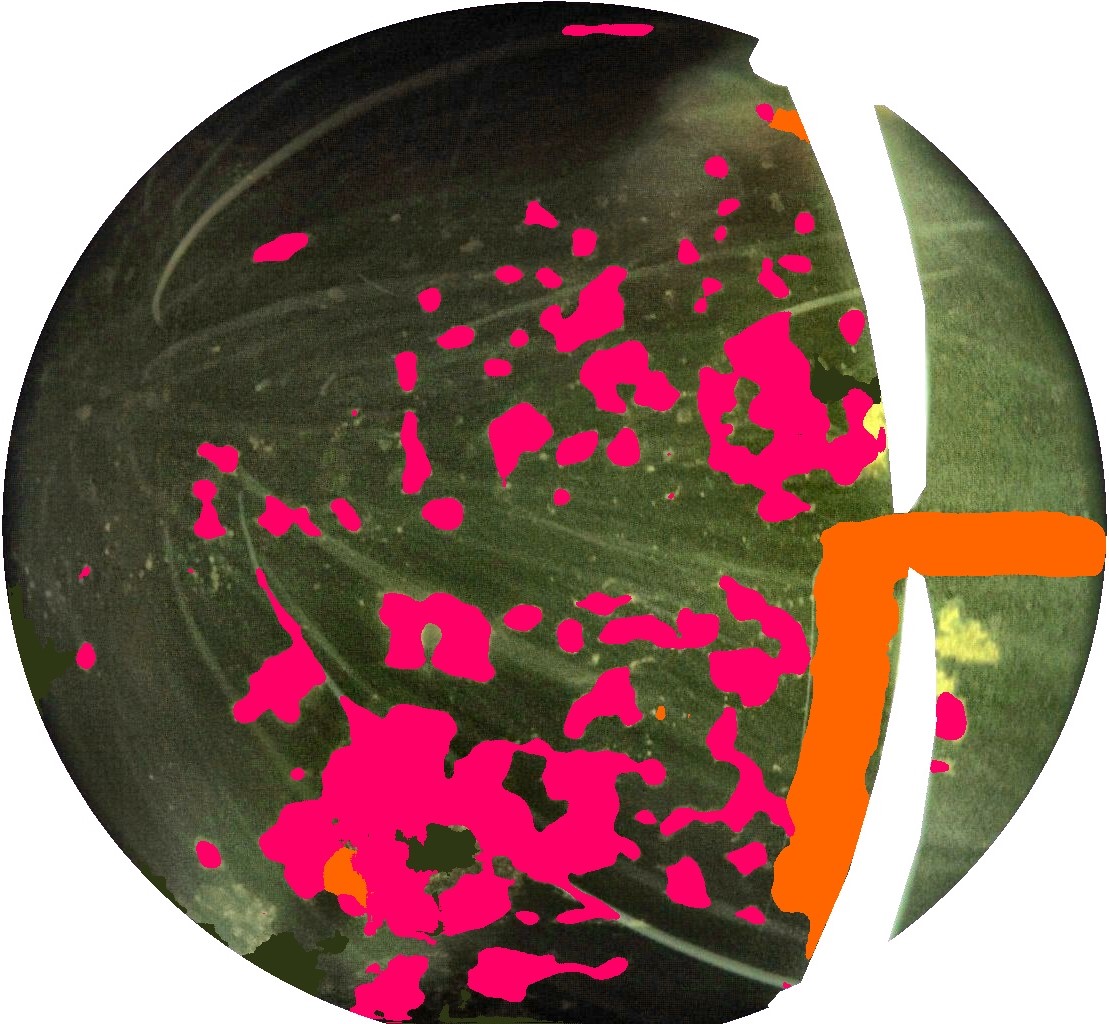}
    \end{minipage}  \\
\vspace{.1in}

    \begin{minipage}{.24\textwidth}
      \vspace{-2mm}
      \includegraphics[width=\linewidth]{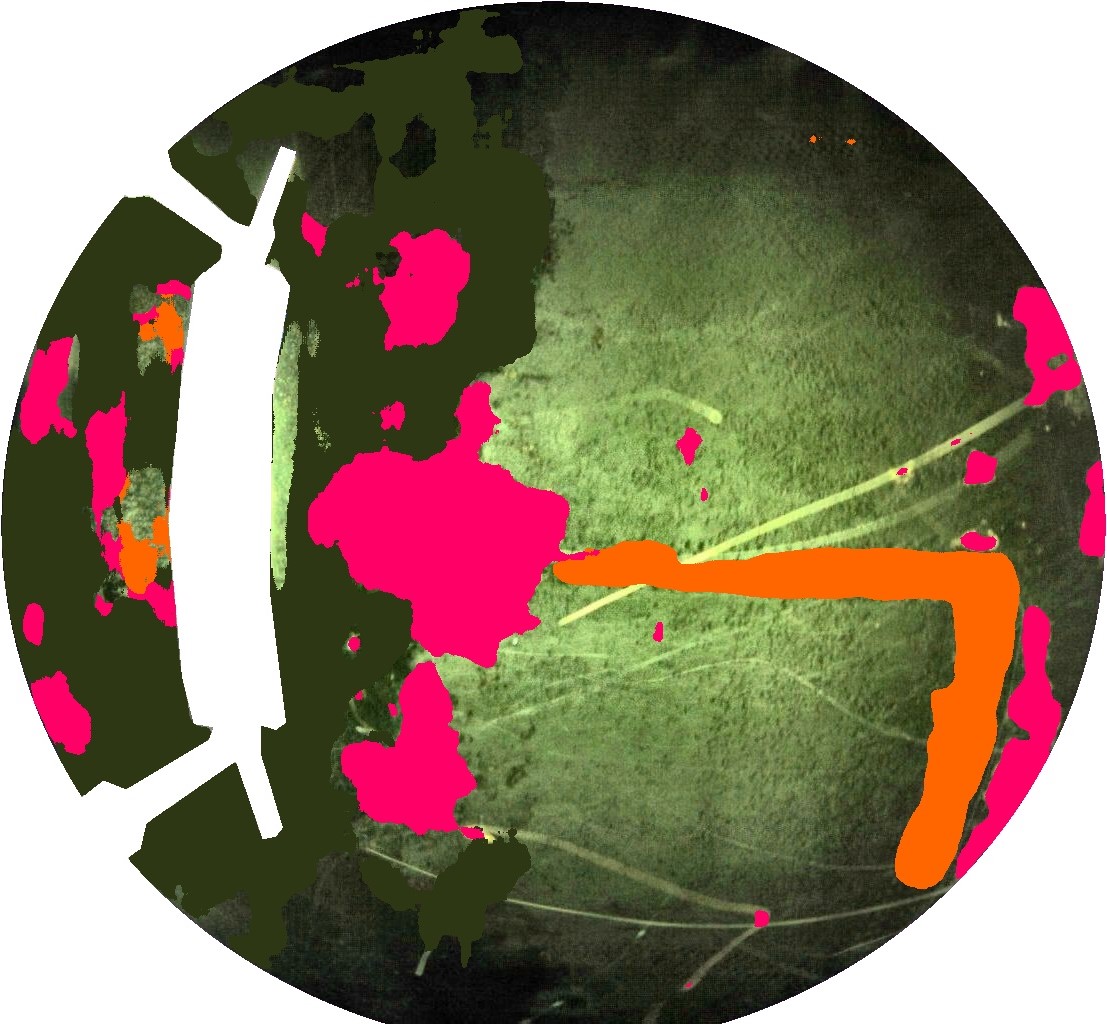}
    \end{minipage}
    \begin{minipage}{.24\textwidth}
      \vspace{-2mm}
      \includegraphics[width=\linewidth]{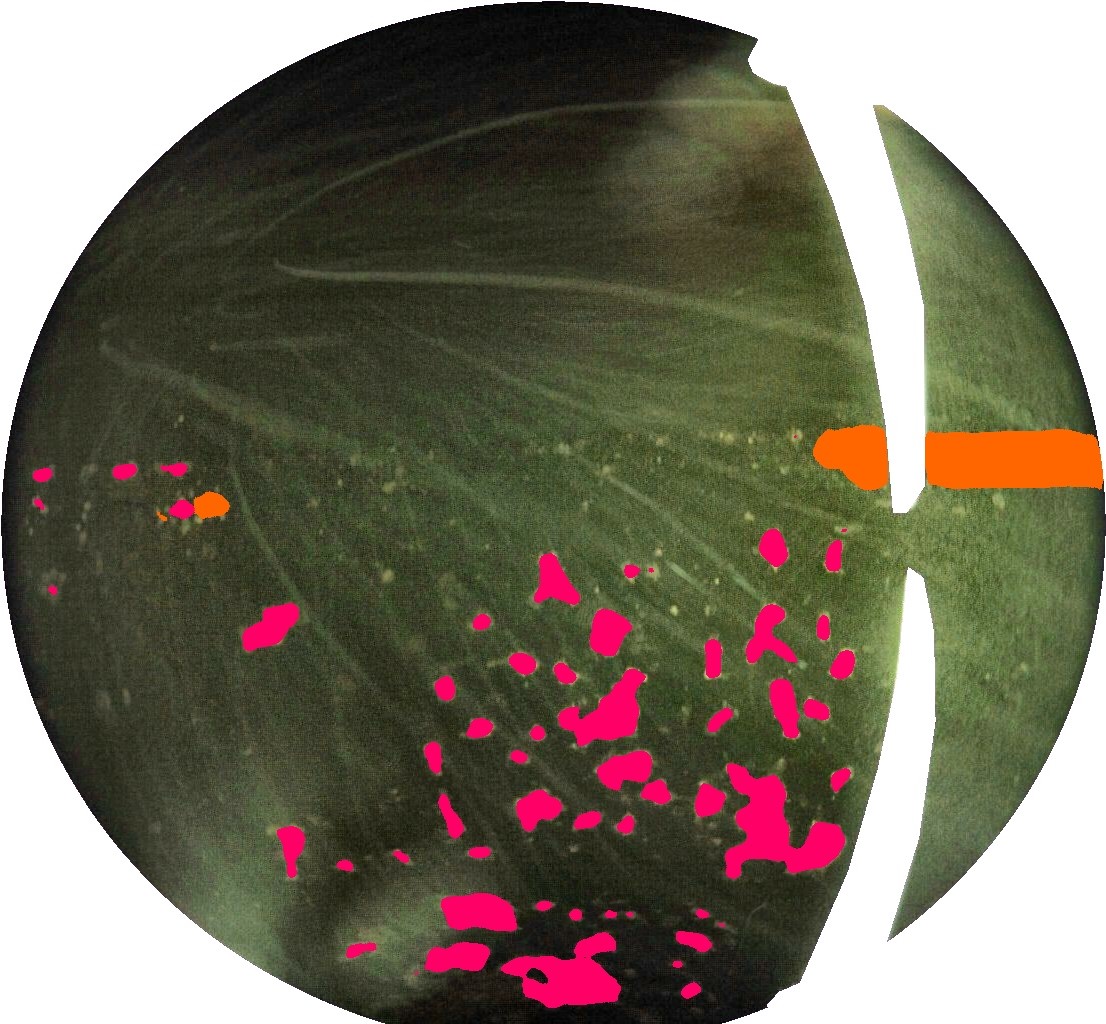}
    \end{minipage}
    \begin{minipage}{.24\textwidth}
      \vspace{-2mm}
      \includegraphics[width=\linewidth]{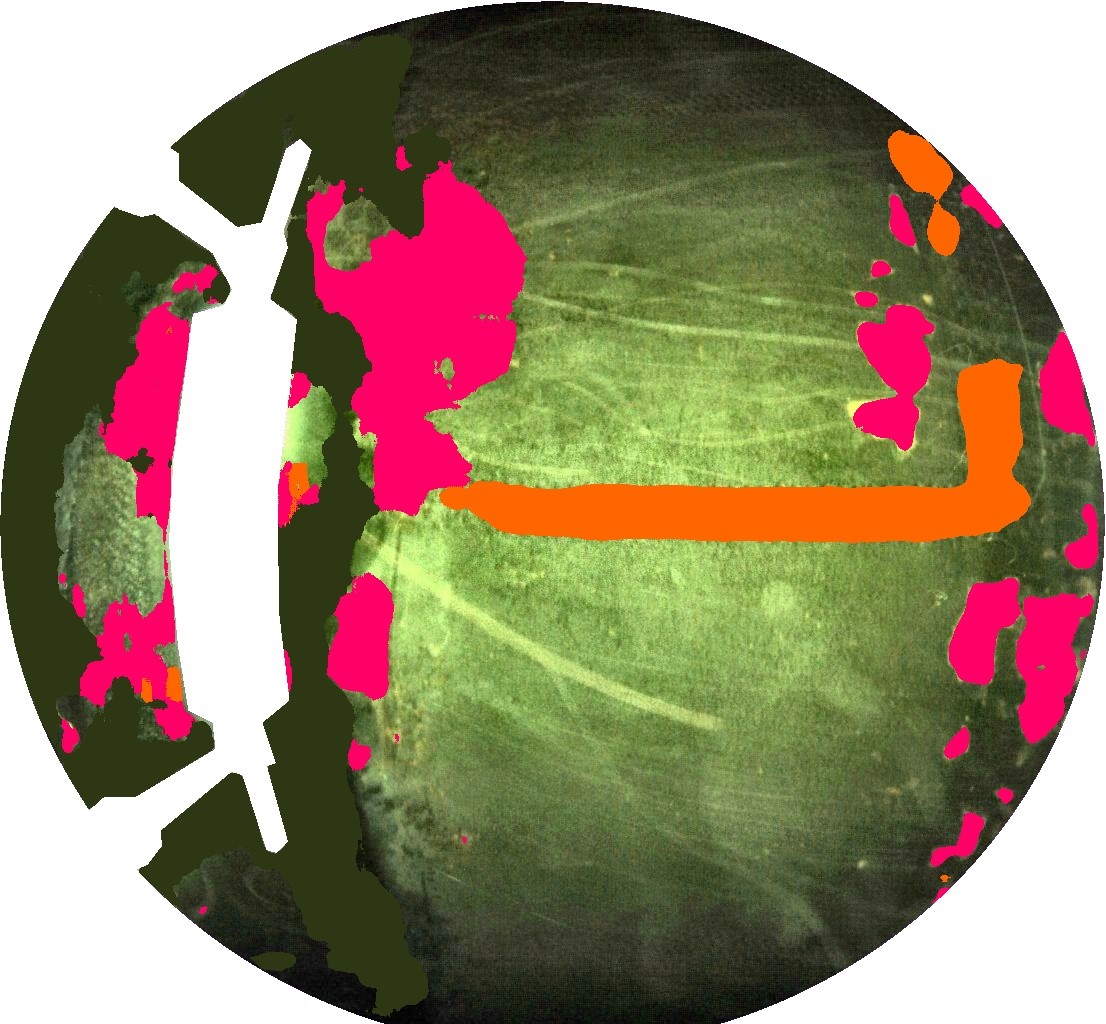}
    \end{minipage}
     \begin{minipage}{.24\textwidth}
      \vspace{-2mm}
      \includegraphics[width=\linewidth]{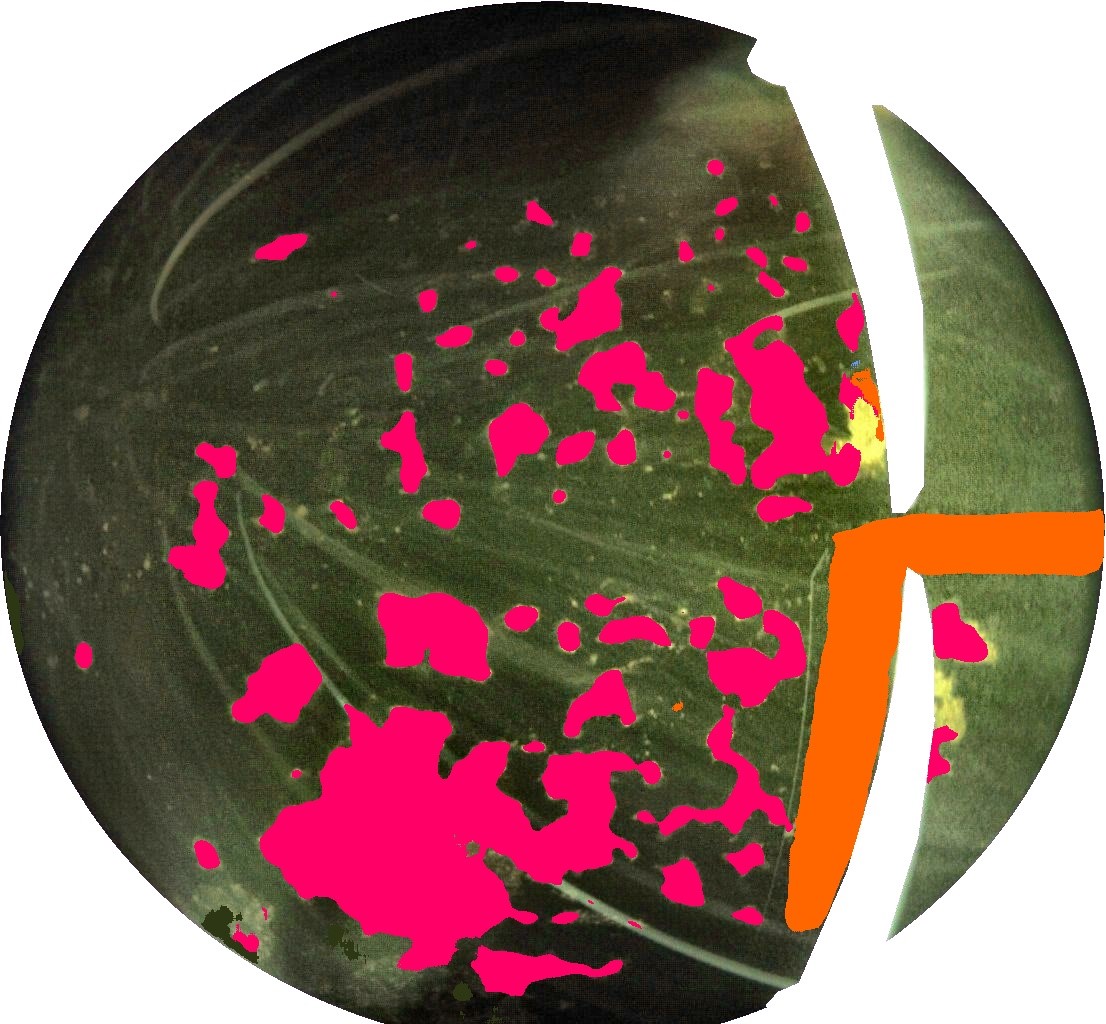}
    \end{minipage}  \\
\vspace{.1in}

    \begin{minipage}{.24\textwidth}
      \vspace{-2mm}
      \includegraphics[width=\linewidth]{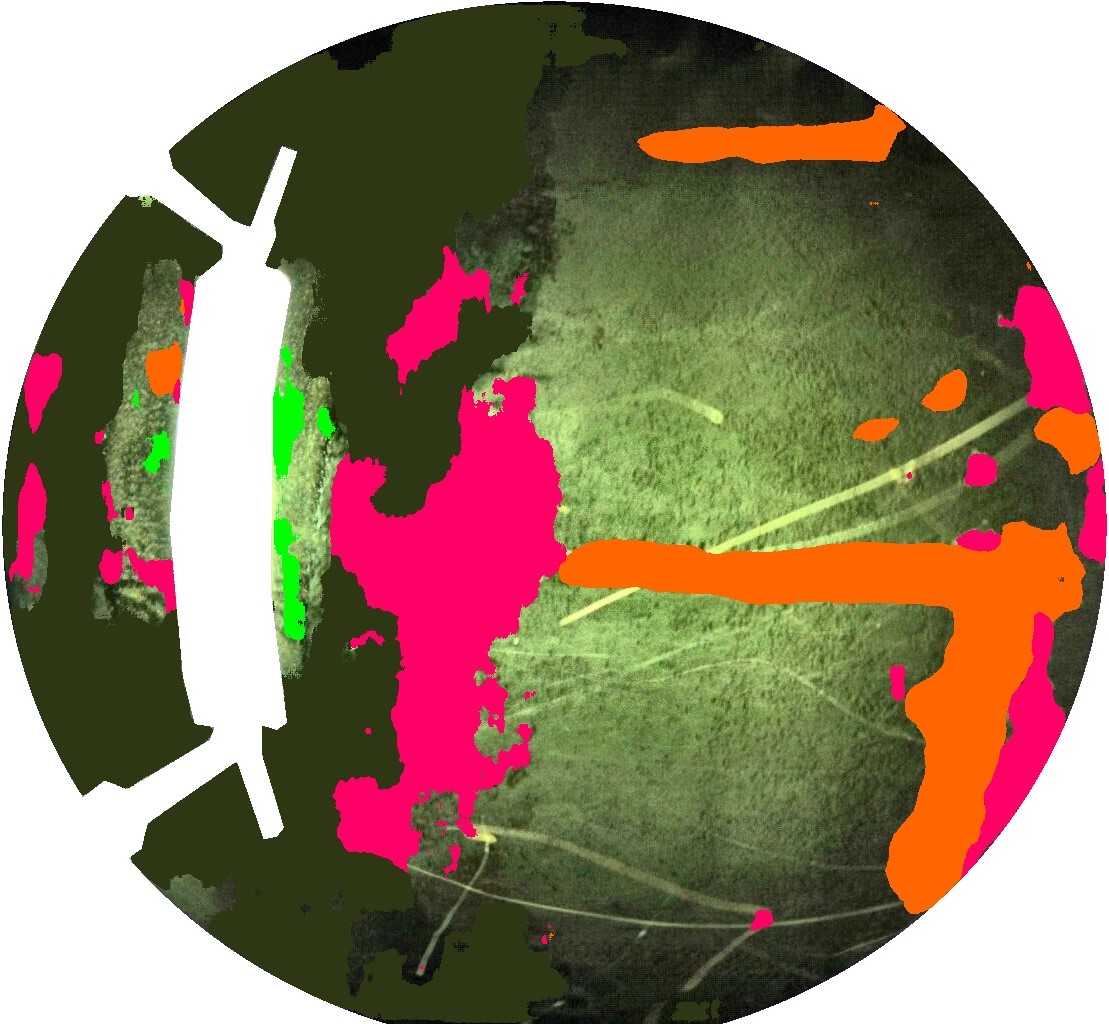}
    \end{minipage}
    \begin{minipage}{.24\textwidth}
      \vspace{-2mm}
      \includegraphics[width=\linewidth]{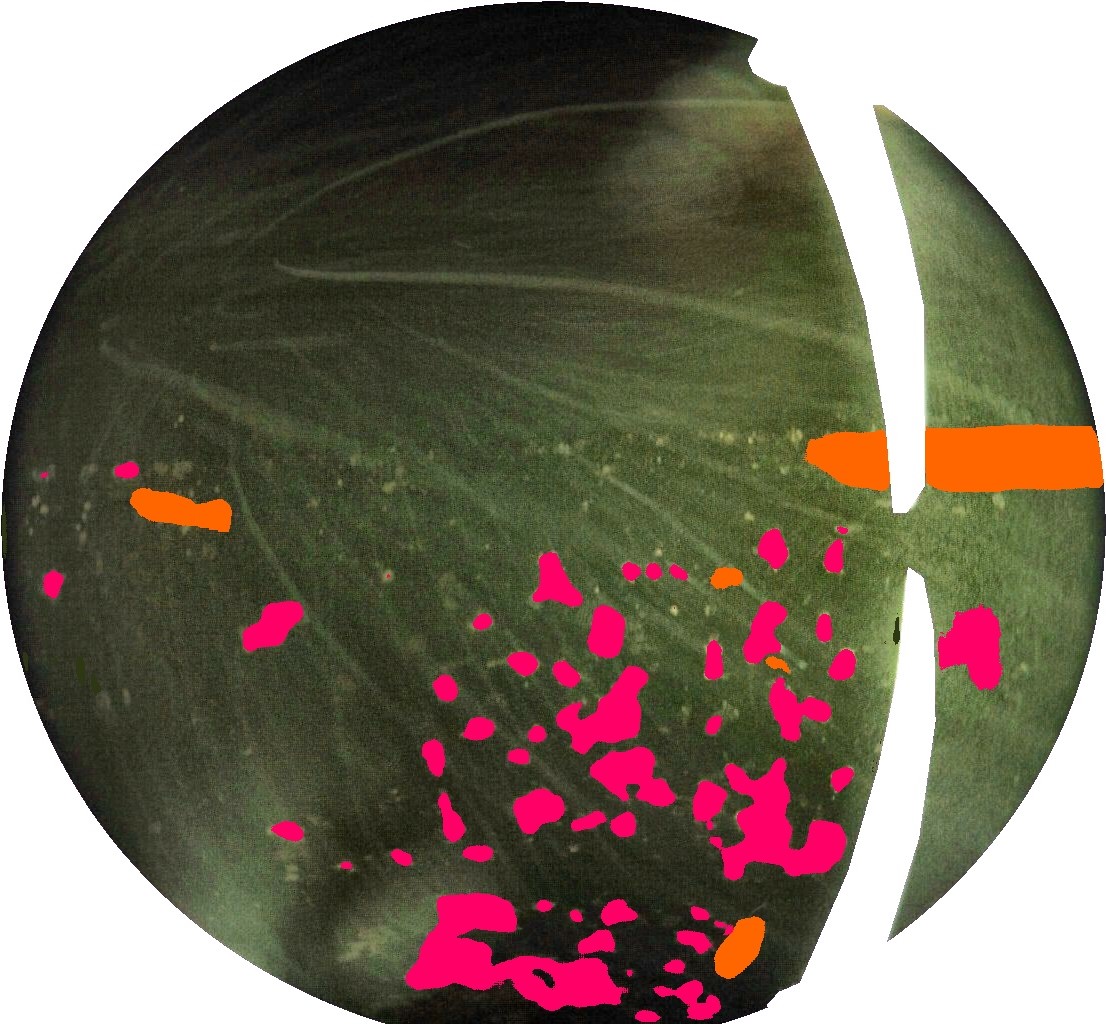}
    \end{minipage}
    \begin{minipage}{.24\textwidth}
      \vspace{-2mm}
      \includegraphics[width=\linewidth]{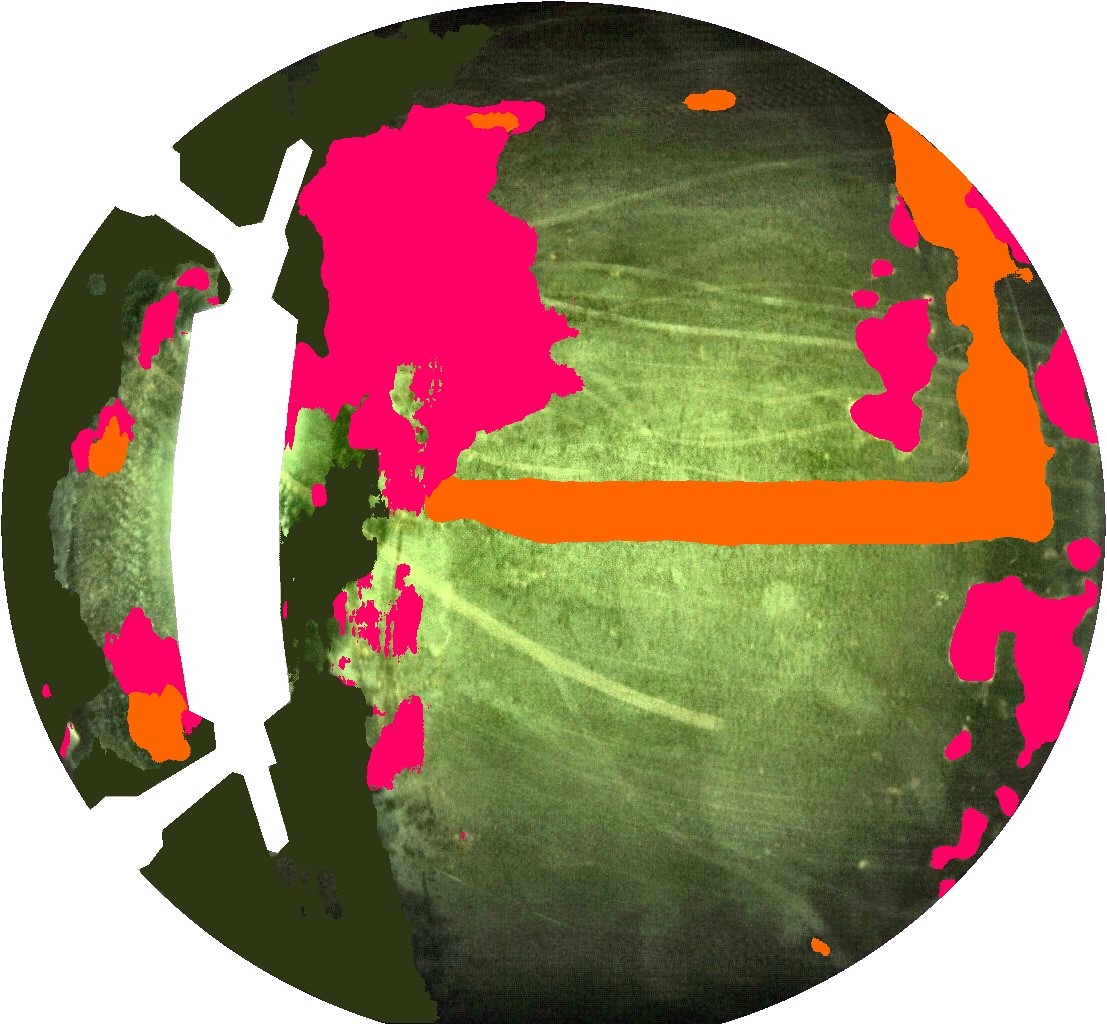}
    \end{minipage}
     \begin{minipage}{.24\textwidth}
      \vspace{-2mm}
      \includegraphics[width=\linewidth]{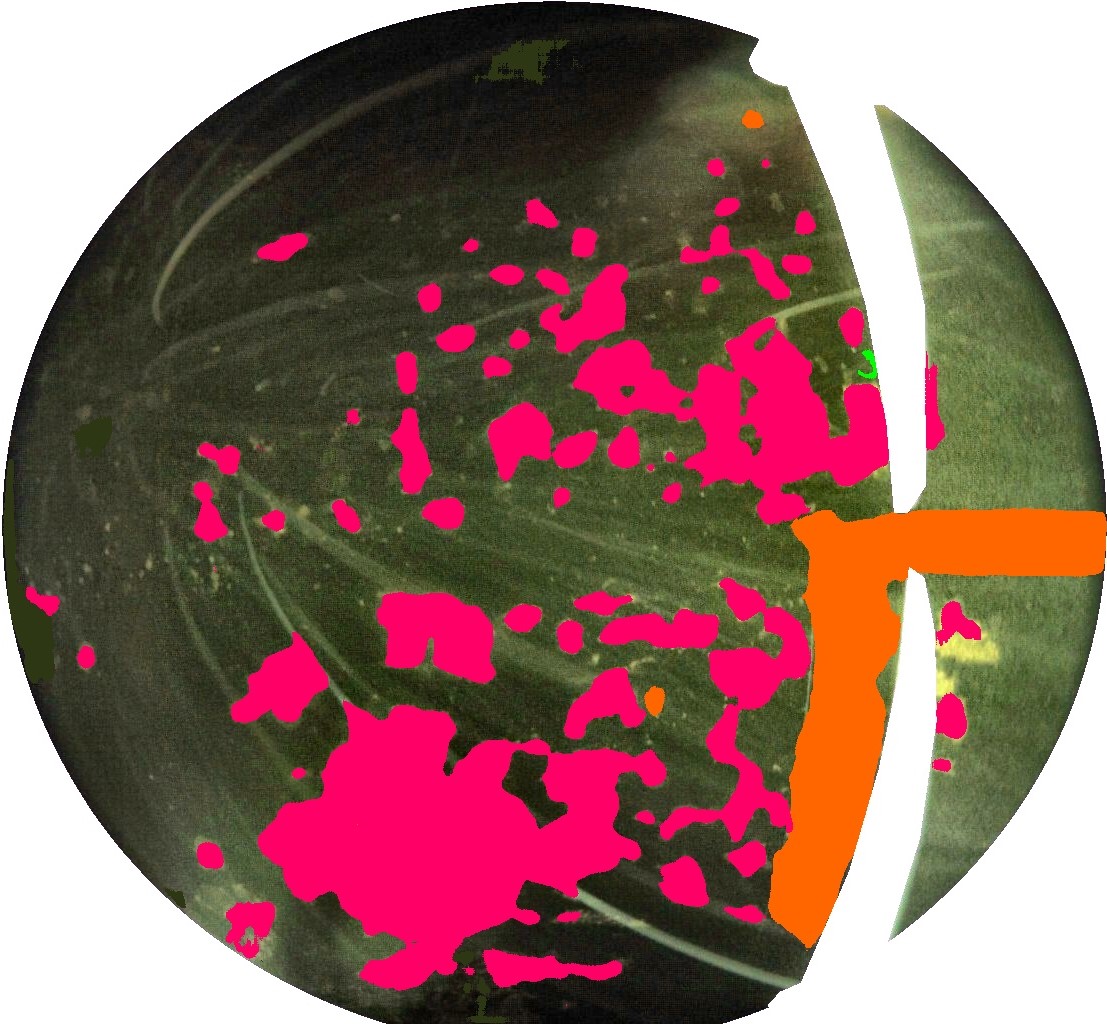}
    \end{minipage}  \\
\vspace{.1in}

    \begin{minipage}{.24\textwidth}
      \vspace{-2mm}
      \includegraphics[width=\linewidth]{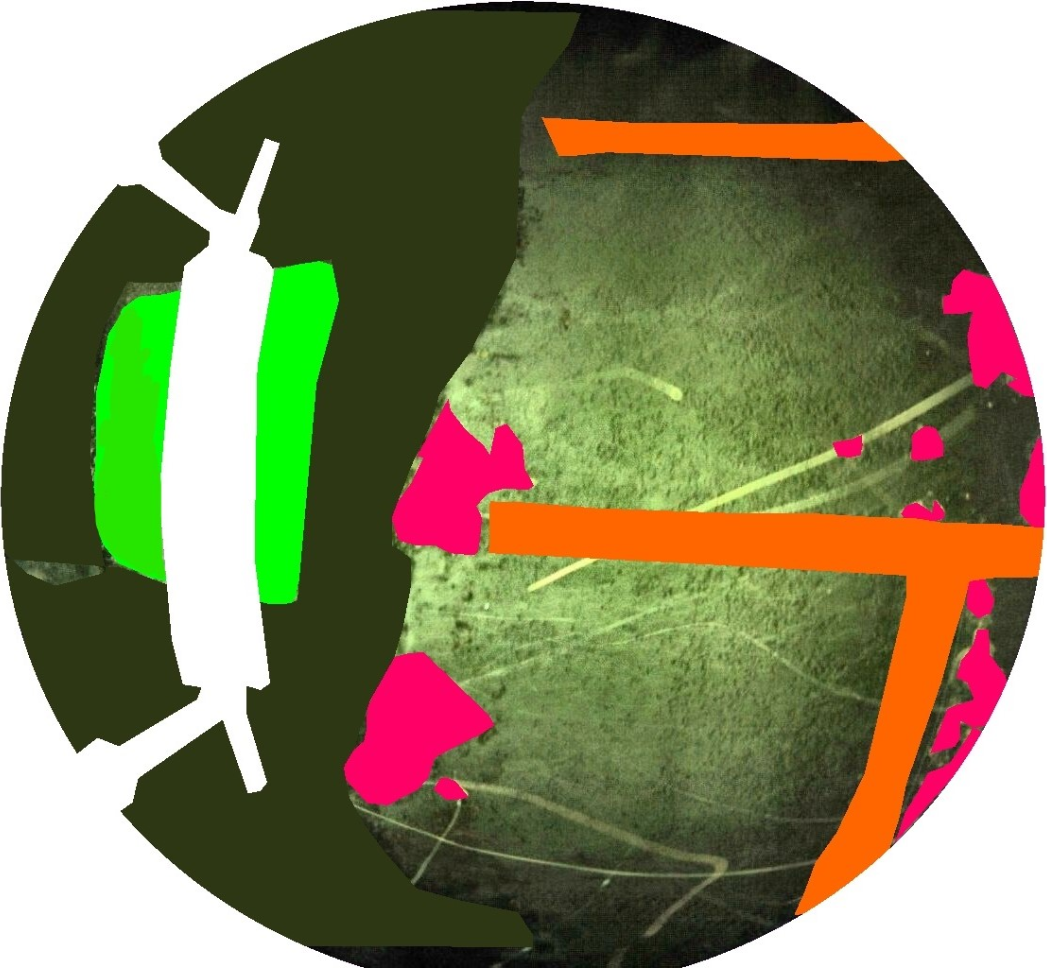}
    \end{minipage}
    \begin{minipage}{.24\textwidth}
      \vspace{-2mm}
      \includegraphics[width=\linewidth]{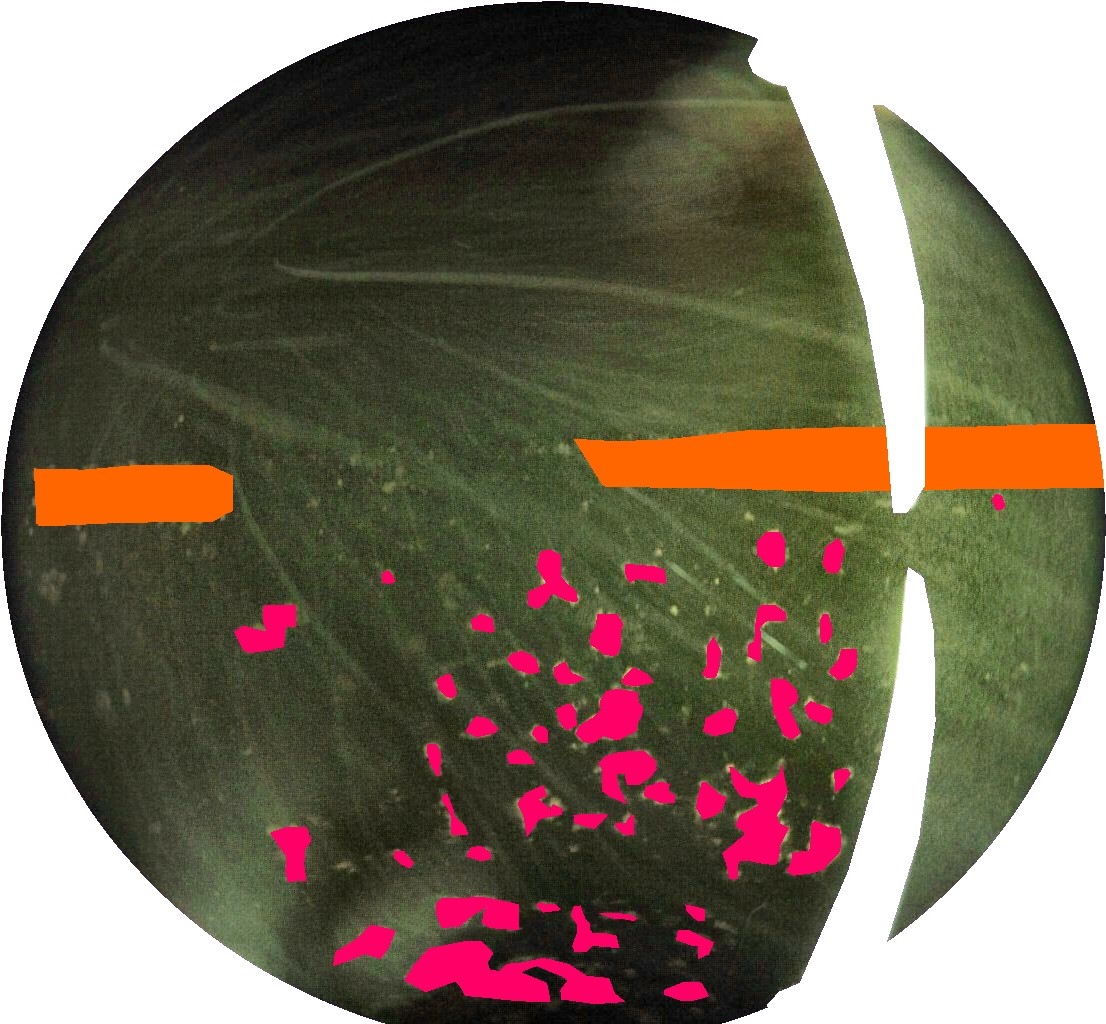}
    \end{minipage}
    \begin{minipage}{.24\textwidth}
      \vspace{-2mm}
      \includegraphics[width=\linewidth]{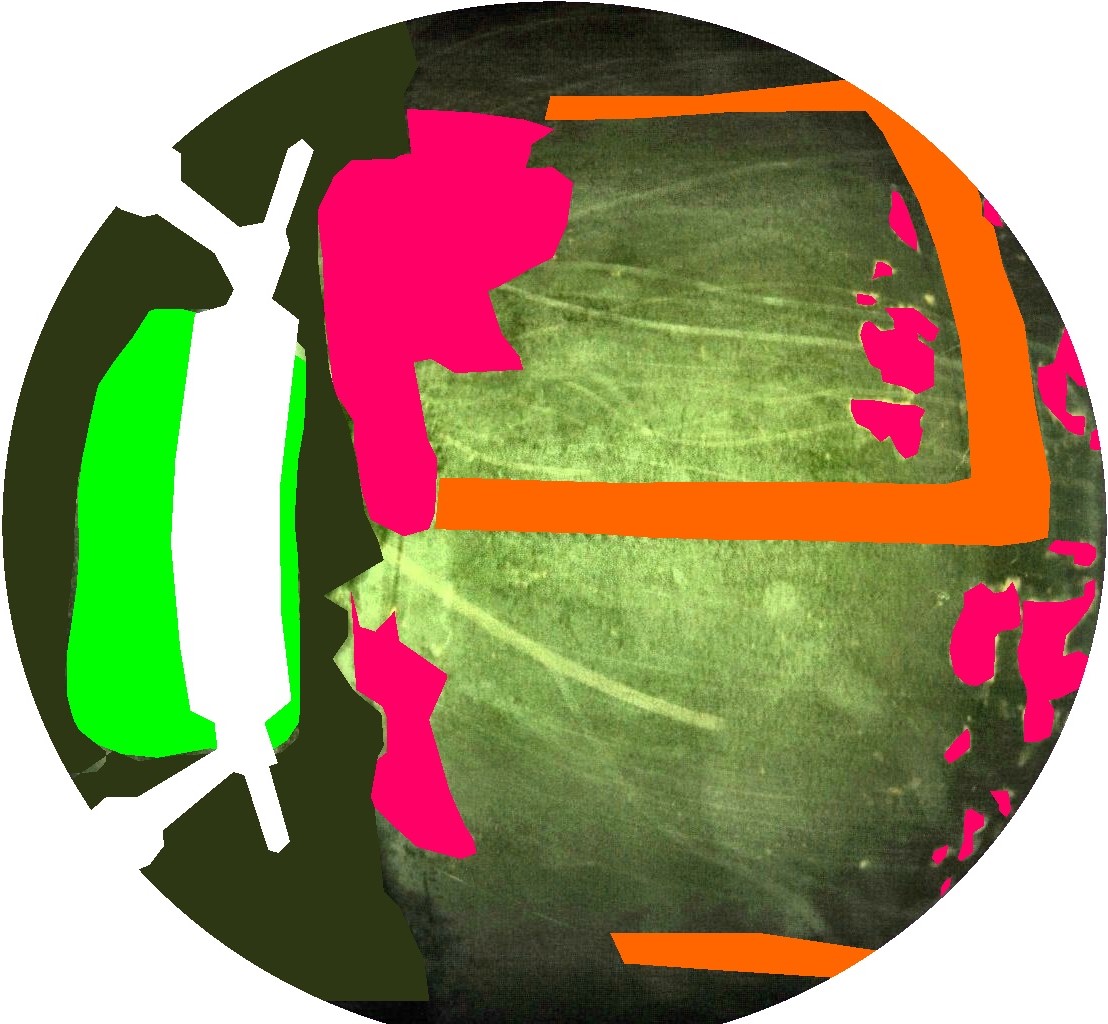}
    \end{minipage}
     \begin{minipage}{.24\textwidth}
      \vspace{-2mm}
      \includegraphics[width=\linewidth]{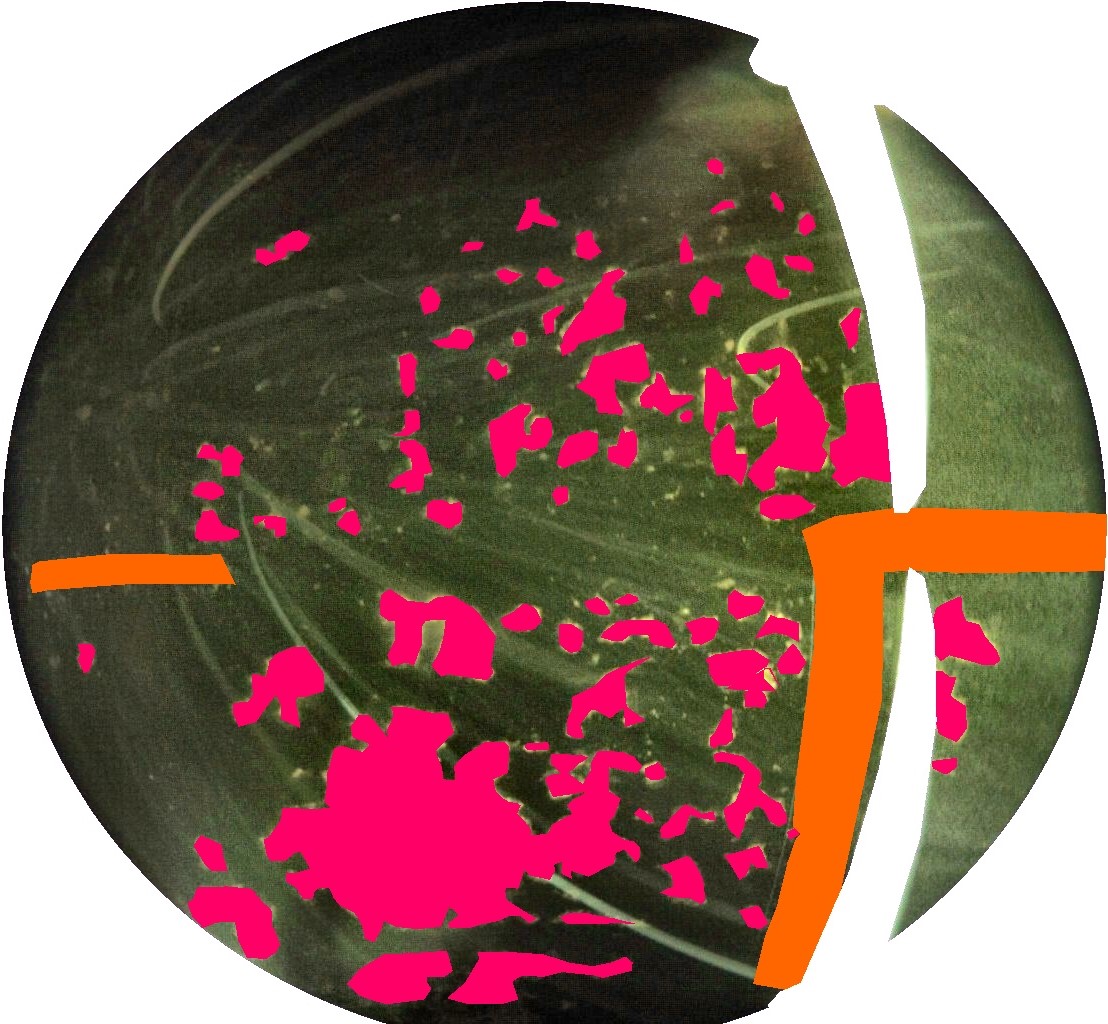}
    \end{minipage} 
\vspace{.1in}

\caption{Qualitative visualization showing the performance of the network with different loss functions with other parameters kept the same. From left to right: four sample images segmented by the network trained by different loss functions. 
First row: softmax cross-entropy; second row: weighted softmax cross-entropy; third row: focal loss ($\gamma = 2.0$); fourth row: weighted focal loss ($\gamma = 2.0$); fifth row: ground truth. 
Overall, weighted focal loss performs the best.  }
\label{fig:qualitative_result}
\vspace{5mm}
\end{figure*}

Tab.~\ref{tab:quant_result} shows the quantitative results while Fig.~\ref{fig:qualitative_result} shows the qualitative results of different loss functions with fixed class weights based on four metrics. 
We use softmax cross entropy as the base loss to which other losses are compared. 

As seen in  Fig.~\ref{fig:qualitative_result}, row $1$, the base loss fails to preserve the boundaries of corroded regions due to the imbalance problem. 
As a consequence, its sensitivity values are much smaller than its specificity values.  



Incorporating class weights to the base loss function can significantly improve the sensitivity value from, $67.5$ to $75.2$, increase DSC and lower total error values. 
The visualization in Fig.~\ref{tab:quant_result}, row $2$ shows that the weighted softmax cross entropy loss helps preserving the boundaries better than the base loss at the expense of having more false positives. 
This is because class weights help balancing the contribution of classes to the total loss giving more training signal from the positive samples.  

On the other hand, the focal loss is shown to be superior than the base loss in specificity and DSC values. 
However, it also fails to preserve the boundaries of corroded regions as in the base loss case. 
Its sensitivity values do not surpass those of the base loss. 
The visualization in Fig.~\ref{tab:quant_result}, row $3$ demonstrates that the focal loss tends to have less false positives than the weighted softmax entropy but have more false negatives resulting in higher specificity but lower sensitivity. 
This suggests that, in this case, the focal loss tends to consider false positive samples \textit{harder} than false negative samples. 
This can be explained by the fact that there are dust noise as well as reflective regions on images that look like corrosion. 
During the training, the network may output low confidence on classifying them as non-corroded. 
Another attribute to low sensitivity can come from wrong annotations, especially in the image regions that the labeler is uncertant about the correct label, as we discuss in Sec.~\ref{sec:dataset}. 

Taking the advantages of weighted loss and the focal loss into account, the weighted focal loss performs better than other methods. 
Indeed, it has a draw with the focal loss on DSC values while yielding higher sensitivity values and lower total error values. 
In our experiments, the weighted focal loss and focal loss performs the best with $\gamma=2.0$.   
Fig.~\ref{tab:quant_result}, row $3$ shows that not only corrosion but wet region and rivet regions are cleaner and more complete in weighted loss function results. 

Comparison on the whole video images can be found at \href{https://drive.google.com/a/seas.upenn.edu/file/d/1PqxkEGlICBazq8jUcMzgHHjdgCUbURRK/view?usp=sharing}{this source}.

\subsection{Effects of Class Weights}

In attempts to investigate the effect of class weights on the performance of loss functions, we conduct an experiment with W-SCE and W-Focal losses in which we adjust the weight of class $\textbf{corroded}$ while keeping the remaining class weights and other settings the same. 
Let ($w_c$) be the weight value of corroded pixel after being normalized by $5$ (thus, $w_c=1$ corresponds to results shown in Tab.~\ref{tab:quant_result}).
We train the networks as before with W-SCE and W-Focal losses using three additional values of $w_c$ which are $1/4, 4, 8$ and report the results in Tab.~\ref{tab:quant_result_weights}. 

\begin{table}[t]
\centering
\begin{tabular}{|c|c|c|c|c|c|}
\hline
Loss    & $w_c$ & DSC  & Sensitivity  & Specificity & Total Error  \\ \hline
\multirow{4}{*}{} 
   & $1/4$ & $50.0$ & $67.0$  & $97.6$ & $2.8$ \\ \cline{2-6} 
W- & $1$ & $51.1$ & $75.2 $ & $97.0$ & $2.5$ \\ \cline{2-6} 
   SCE & $4$ & $48.4$  &$82.0$  & $95.3$ & $2.1$\\ \cline{2-6} 
& $8$ & $44.5$  &$83.4$  & $95.3$ & $2.2$\\ \hline
\multirow{4}{*}{} 
   & $1/4$ & $47.4$ & $45.0$  & $99.6$ & $4.8$ \\ \cline{2-6} 
W- & $1$ & $52.1$ & $71.6$ & $97.7$ & $2.2$ \\ \cline{2-6} 
   Focal & $4$ & $47.7$  &$81.4$  & $97.7$ & $1.9$\\ \cline{2-6} 
& $8$ & $45.2$  &$83.3$  & $95.2$ & $2.5$\\ \hline
\end{tabular}
\caption{Performance with different class weights ($\%$)\\
DSC, Sensitivity, Specificity: higher is better. Total Error: smaller is better}
\label{tab:quant_result_weights}
\vspace{-3mm}
\end{table}

The results in Tab.~\ref{tab:quant_result_weights} demonstrate that the class weights largely effect the metrics scores. 
Higher weight for corroded class results in higher sensitivity values, but smaller specificity values since the model will favor the corroded class for the sake of minimizing the loss. 
As a result, DSC and total error values fluctuate around an $'optimal'$ value for each metric. 
For DSC, they are the highest with $w_c=1$. 
For total error, they are optimal with $w_c = 4$. 

These observations repeat suggestion that the loss function is highly dependent to the application and introducing proper class weights can help achieve an optimal solution for a specific metric that we choose. 





%% file: tex/conclusion.tex
\section{Conclusions}
\label{sec:conclusions}
This work attempts to offer an automated solution for safer, faster, cost-efficient and objective infrastructure inspection with a focus on penstocks. 
We propose a data-efficient, data-driven image segmentation method using a fully convolution neural network that can detect highly non-homogeneous objects under low-light and high-noise conditions in real time. 
Our method can be seamlessly combined with other MAV planning algorithms to provide a completely automated and real-time inspection tool to replace humans in labor-intensive and dangerous tasks. 
Our analysis on different loss functions can provide hints to general image segmentation problems with class imbalance. 
The experimental results obtained with the dataset collected at Center Hill Dam, TN demonstrate that the focal loss, in combination with a proper set of class weights yields better segmentation results than the commonly used softmax cross entropy loss. 
One limitation of the focal loss and weighted focal loss is that their outputs tend to vary at different testing times. This can be addressed in a future work.


%% file: tex/appendix.tex
\section{Appendix}
\begin{figure*}[t]
  \vspace{3mm}
   \centering
   \includegraphics[width=\linewidth]{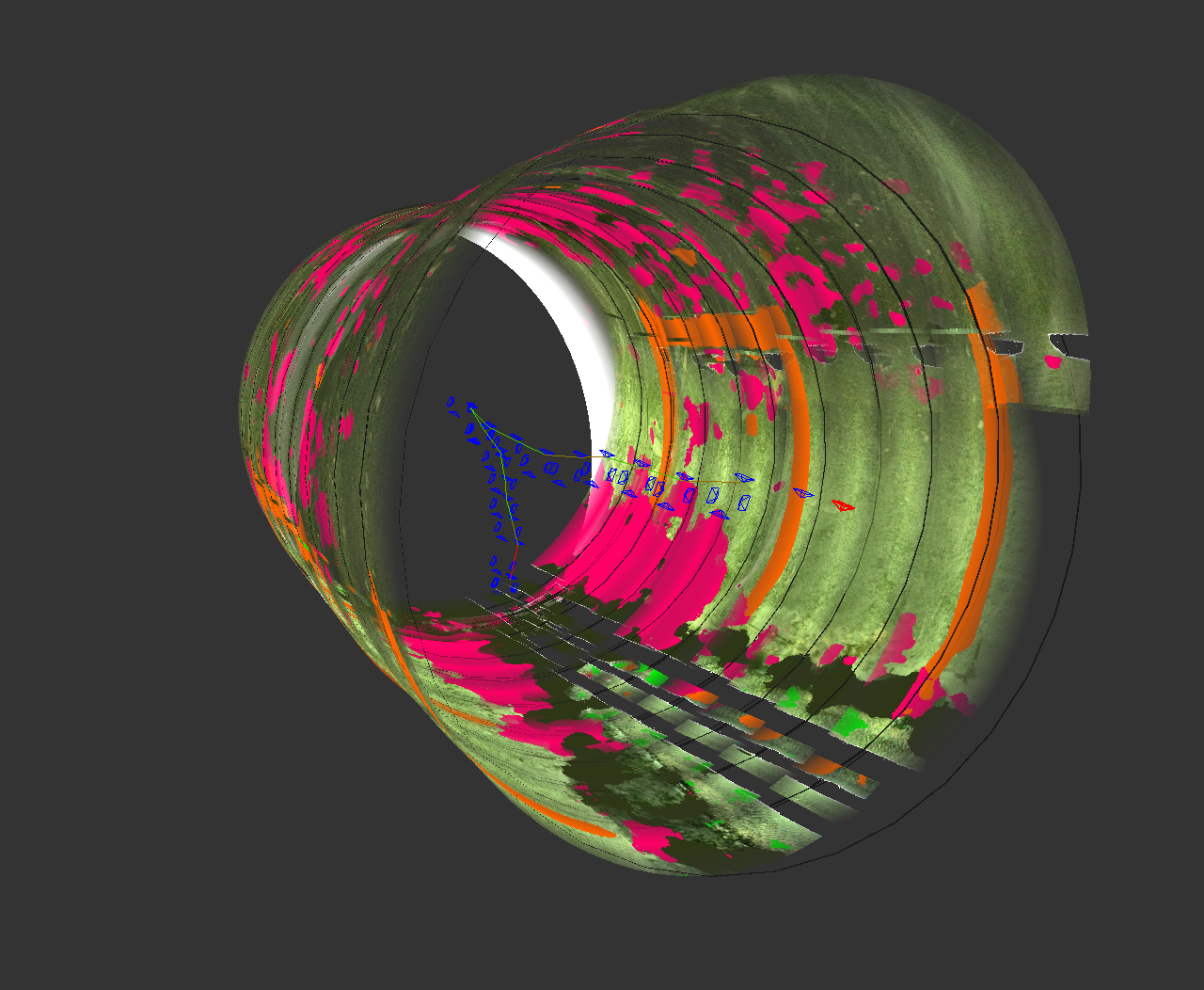}
    \caption{3D volume rendering of the penstock. Pink: corroded, Orange: rivet, Green: water, Pale: coating, Blue: MAV's trajectory}
    \label{fig:intro_image}
    \vspace{-3mm}
\end{figure*}